\documentclass[sn-mathphys]{sn-jnl}

\jyear{2022}%

\theoremstyle{thmstyleone}%
\theoremstyle{thmstyletwo}%

\theoremstyle{thmstylethree}%

\begin{document}

\title[SPDGAN: A Generative Adversarial Network based on SPD Manifold Learning for Automatic Image Colorization]{SPDGAN: A Generative Adversarial Network based on SPD Manifold Learning for Automatic Image Colorization}


\author*[1]{\fnm{Youssef} \sur{Mourchid}}\email{ymourchid@cesi.fr}

\author[2]{\fnm{Marc} \sur{Donias}}\email{marc.donias@ims-bordeaux.fr}

\author[2]{\fnm{Yannick} \sur{Berthoumieu}}\email{yannick.berthoumieu@ims-bordeaux.fr}

\author[2]{\fnm{Mohamed} \sur{Najim(LiveFIEEE)}}\email{mohamed.najim@u-bordeaux.fr}

\affil*[1]{\orgdiv{CESI LINEACT}, \orgname{UR 7527}, \orgaddress{ \city{Dijon}, \postcode{21800}, \country{France}}}

\affil[2]{\orgdiv{Universite de Bordeaux}, \orgname{ Bordeaux INP}, \orgaddress{\street{CNRS, IMS, UMR 5218}, \city{Talence}, \postcode{33400}, \country{France}}}


\abstract{This paper addresses the automatic colorization problem, which converts a gray-scale image to a colorized one. Recent deep-learning approaches can colorize automatically grayscale
images. However, when it comes to  different scenes
which contain distinct color styles, it is difficult to accurately capture the color characteristics. In this work, we propose a fully automatic colorization approach based on Symmetric Positive Definite (SPD) Manifold Learning with a generative adversarial network (SPDGAN) that improves the quality of the colorization results. Our SPDGAN model establishes an adversarial game between two discriminators and a generator. The latter is based on ResNet architecture with few alterations. Its goal is to generate fake colorized images without losing color information across layers through residual connections. Then, we employ two discriminators from different domains. The first one is devoted to the image pixel domain, while the second one is to the Riemann manifold domain which helps to avoid color misalignment. Extensive experiments are conducted on the
Places365 and COCO-stuff databases to test the effect of each component of our SPDGAN. In addition, quantitative and qualitative comparisons with state-of-the-art methods demonstrate the effectiveness of our model by achieving more realistic colorized images with less artifacts visually, and good results of PSNR, SSIM, and FID values.
}


\maketitle

\section{Introduction}\label{sec:intro}

Image colorization goal is to convert an image from grayscale to another color space so that the image is colorized. This task has attracted a lot of attention from researchers due to its wide range of applications. It is still an active topic in the field of image processing research since it has so many useful applications\cite{mourchid2016image,benallal2022new,mourchid2021automatic,mourchid2023mr}, like coloring old black-and-white pictures, creating artwork, etc.
Hence, the problem of image colorization is stated as ill-posed, the majority of the computer graphics work done for image colorization can be mainly divided into four categories: scribble-based colorization \cite{levin2004colorization,huang2005adaptive,luan2007natural}, example-based colorization \cite{welsh2002transferring,chia2011semantic,liu2008intrinsic}, deep learning-based colorization \cite{deshpande2015learning,cheng2015deep,zhang2016colorful,larsson2016learning,zhang2017real,xia2018scene} and GAN-based methods \cite{isola2017image,nazeri2018image,vitoria2020chromagan,mourchid2020dual,mourchid2021automatic}. The scribble-based methods typically demand extra work from the user to make significant scribbles on the grayscale images. These methods take a long time to colorize a grayscale image with fine-scale structures, though. Such work \cite{welsh2002transferring} presented an example-based strategy to reduce manual user effort, which was later enhanced by \cite{charpiat2008automatic}. Authors use color scribbles, which transfer color using a second color reference image, to remove the burden of annotated images.

Nevertheless, it is hard for users to find a suitable reference image. Other works \cite{chia2011semantic,liu2008intrinsic} simplify this problem using the image data on the Internet and for selecting suitable reference images they propose filtering schemes. However, both approaches have some limitations, and their performance highly depends on the selected reference images. To model large-scale data,  deep learning techniques are suitable to achieve amazing success. To some extent (e.g., \cite{he2015delving}), they even outperform human beings through their powerful learning ability, and they have been very performed for various applications such as image colorization tasks. A wide range of works \cite{deshpande2015learning,cheng2015deep,zhang2016colorful,larsson2016learning,zhang2017real} have used various convolutional network architectures with different loss functions to resolve the image colorization problem.

A new generation of learning structures known as Generative Adversarial Networks was recently introduced by Goodfellow \emph{et al.} \cite{goodfellow2014generative}. It was frequently used to create artificial data with identical statistics to the real one. GANs employ a clever training method that makes two neural networks in competition: a generator and a discriminator. The generator tries to generate as many realistic data points as possible to fool the discriminator while the latter tries to distinguish between generated fake data points and real ones. Recent works in colorization have shown that GANs can improve the perceptual quality of generated images compared to other deep-learning-based models. For this reason, these architectures have inspired us to explore their potential application in our work.

The importance of learning the textured style content of images from Symmetric Positive Definite (SPD) matrices (ex: covariance matrices) has been demonstrated in other works, such as style transfer \cite{gatys2016image} or super-resolution \cite{johnson2016perceptual,balas2006texture}, to improve the quality of the generator outputs. Traditional GANs architectures used in the previous works capture only first-order statistics, because they are based on convolutional neural networks (CNNs), while it was considered that second-order statistics are better regional descriptors compared to first-order statistics. In this context, we introduce Symmetric Positive Definite (SPD) matrices which can be employed as second-order descriptors.

SPD matrices have attracted much attention in a variety of applications. They provide powerful statistical representations for images in visual recognition. Such applications employ covariance matrices to detect pedestrian, or joint covariance descriptors for action recognition. The non-Euclidean data structure of these matrices, which underlies a Riemannian manifold, makes it difficult to interpolate or restore them for their classification. So, applying directly Euclidean geometry to SPD matrices may produce undesirable effects. Several learning approaches were proposed to address this problem by mapping SPD matrices via a tangent space approximation or kernel Hilbert spaces \cite{zhen2021flow}. With the success of deep learning approaches, Huang \emph{et al. }\cite{huang2017riemannian} proposed a deep Riemannian neural network architecture to perform non-linear computations of the SPD matrices with effective backpropagation training algorithms. This network takes as an input an SPD matrix and then preserves the SPD structure across layers, and offers a possibility of SPD matrix non-linear learning by a new backpropagation technique.

By using generative adversarial networks (GANs) architecture and SPD manifold learning, we introduce a novel automatic image colorization method in this work that overcomes the drawbacks of earlier conventional GAN-based approaches. Our proposed architecture employs two discriminators, the first one is for the pixel level, while the second one is on the Riemannian manifold. In summary, the main contributions of this work are:

\begin{itemize}
    \item a SPDGAN model for automatic colorization which uses two discriminators to ensure the restitution of fine structures of the produced images. The first one is employed at the pixel level while the second one is for the image feature level.
    
    \item An encoder-decoder network based on ResNet architecture which resolves the degradation caused by the increase of the number of layers in other CNN models.
    
    \item A new direction of SPD matrix in the context of image colorization is opened by investigating the Riemannian network architecture as a feature discriminator.
    
    \item A non-adversarial color loss for the comparison of contrast, brightness, and major colors between colorized images and ground truth ones.
    
\end{itemize}

This paper is organized as follows: in the following sections, we first review related work in Section \ref{sec:related}, then we introduce in Section \ref{sec:Method} our SPDGAN model. We discuss how well SPDGAN performs in Section \ref{sec:exp}, Finally, we come to conclusion in Section \ref{sec:conclu}.

\section{Related Works} \label{sec:related}

This section offers a brief overview of the previous colorization methods. Moreover, since our proposed approach is based on GAN architecture, we also provide a brief explanation of its basic theory.

\subsection{Scribble-based methods:} 

The scribble-based methods introduced by Levin \emph{et al.} \cite{levin2004colorization}, require substantial efforts from the user to add some scribbles to the target image and these known colors will be propagated to all image pixels. It is assumed that adjacent pixels with similar luminance should have a similar color. Huang \emph{et al.} \cite{huang2005adaptive} proposed an improved method by adjusting the loss function and processed images by edge detection algorithm, then each segmented region will be processed by colorization algorithm. Another work by Luan \emph{et al.} \cite{luan2007natural} employs texture similarity to reduce user interactions. Since human interaction plays an important role in these processes, colorization highly relies on the user's ability. So, improper initial supplied scribbles will lead to unsatisfactory results. Moreover, for massive images (e.g., films) the colorization process of grayscale images with fine-scale structures is time-consuming.

\subsection{Example-based methods:} Example-based methods require a user,  to provide reference image(s), for transferring the color information from a similar reference image to the target grayscale image. Welsh \emph{et al.} \cite{welsh2002transferring}
proposed a technique to colorize grayscale images by creating a set of sample pixels
then each pixel in the grayscale image will be scanned to find the best matching pixel in
the set and transfer chrominance from the sample pixel to the target pixel. However, finding a suitable reference image becomes a difficult task for the user. Chia \emph{et al.} \cite{chia2011semantic} simplify this problem by proposing a method that allows users to input the semantic label for each object in the scene, then automatically download suitable images from the internet and apply a colorization algorithm. In this line, Liu \emph{et al.} \cite{liu2008intrinsic} requires identical Internet objects for precise per-pixel registration between the grayscale image and the reference images. In practice, it is difficult to obtain manual segmentation cues, due to the multiple complex objects (e.g. building, car, tree, elephant) that the target grayscale image may contain. These techniques share the same limitation and their performance highly depends on the selected reference image(s).

\subsection{Deep learning-based methods:} In recent years, deep learning techniques have achieved amazing success in automatic colorization. By employing a large number of color images for training, the relationship between luminance and chrominance channels is determined. Several techniques rely entirely on learning to produce the colorization result. Deshpande \emph{et al.} \cite{deshpande2015learning} proposed a learning method that considers the colorization problem as a quadratic objective function. To improve the initial colorization results they applied a histogram correction. Cheng \emph{et al.} \cite{cheng2015deep} proposed a method that combines three levels of features by increasing the receptive field and feeding it into a three-layer fully connected neural network that minimizes the L2 loss. In \cite{zhang2016colorful}, Zhang \emph{et al.} suggested training a convolutional neural network architecture to reduce the multinomial cross-entropy loss for color distribution prediction. The network is initially seeded with a classification-based network for optimal colorization results. Authors encouraged the use of unlabeled data for automatic colorization based on self-supervision in \cite{larsson2016learning}. In parallel, Zhang \emph{et al.} \cite{zhang2017real} investigated cross-channel encoders, which are an adaptation of the conventional auto-encoder architecture. Wang \cite{wang2022image} \emph{et al.} present the use of the U-net network for image colorization with some modifications (CU-net) to reduce network depth and improve accuracy. The sigmoid activation function and batch normalization are applied, and the extended convolution is introduced for a larger receptive field.

\begin{figure*}[h!]
\begin{center}
\includegraphics[width=13cm,height=3.1cm]{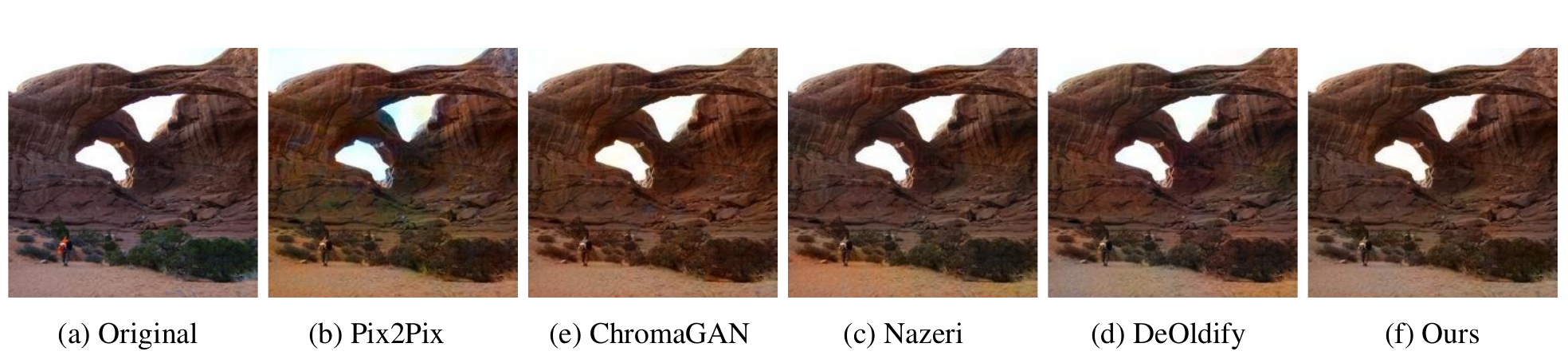}
\caption{Figure is better seen zoomed on the digital version of this document. The limitations of state-of-the-art methods for the colorization process.} 
\label{figure:gan}
\end{center}

\end{figure*}

\subsection{GAN-based methods:} GANs have been recently applied to image colorization by various authors. Isola \emph{et al.} \cite{isola2017image} took advantage of the power of conditional GANs by coupling a DCGAN \cite{radford2015unsupervised} using the U-net architecture for the generator definition and  for PatchGAN architecture the discriminator. Nazeri \emph{et al.} \cite{nazeri2018image} employed a generative network that minimizes an L1 loss in architecture with skip connections (U-Net). Xiao \emph{et al.} \cite{xiao2019single} consider the colorization process as an image-to-image translation based on a CycleGAN model which generates RGB colors by employing high-level semantic identity loss and low-level color loss. Victoria \emph{et al.} \cite{vitoria2020chromagan} propose a model that exploits geometric, perceptual, and semantic features via a self-supervised GAN architecture. By employing the semantic understanding of the real scenes, their proposed architecture can colorize the image  with plausible realistic color rather than just focusing on aesthetic appeal.

In Fig. \ref{figure:gan} we show the common problems of the colorization results of the previous methods \cite{DeOldify,vitoria2020chromagan,isola2017image,nazeri2018image}. The result of Pix2Pix \cite{isola2017image} and ChromaGAN \cite{vitoria2020chromagan} approaches occur a lot of mismatching colors as shown in Fig. \ref{figure:gan}(b)(c). Furthermore, the quality of colorized image in Fig. \ref{figure:gan}(e)(f) has been improved, however, some mismatching colors in some zones still exist. For instance, in Fig. \ref{figure:gan}(e)(f) DeOldify \cite{DeOldify} and Nazeri \cite{nazeri2018image} methods  cannot keep the same color of the ground as the original image. The main reason is, during feature extraction, existing approaches usually lose information details and cannot successfully learn object-level semantics.

\section{Proposed Model}
\label{sec:Method}

\begin{figure*}[htbp]
\begin{center}
\includegraphics[width=13cm,height=7cm]{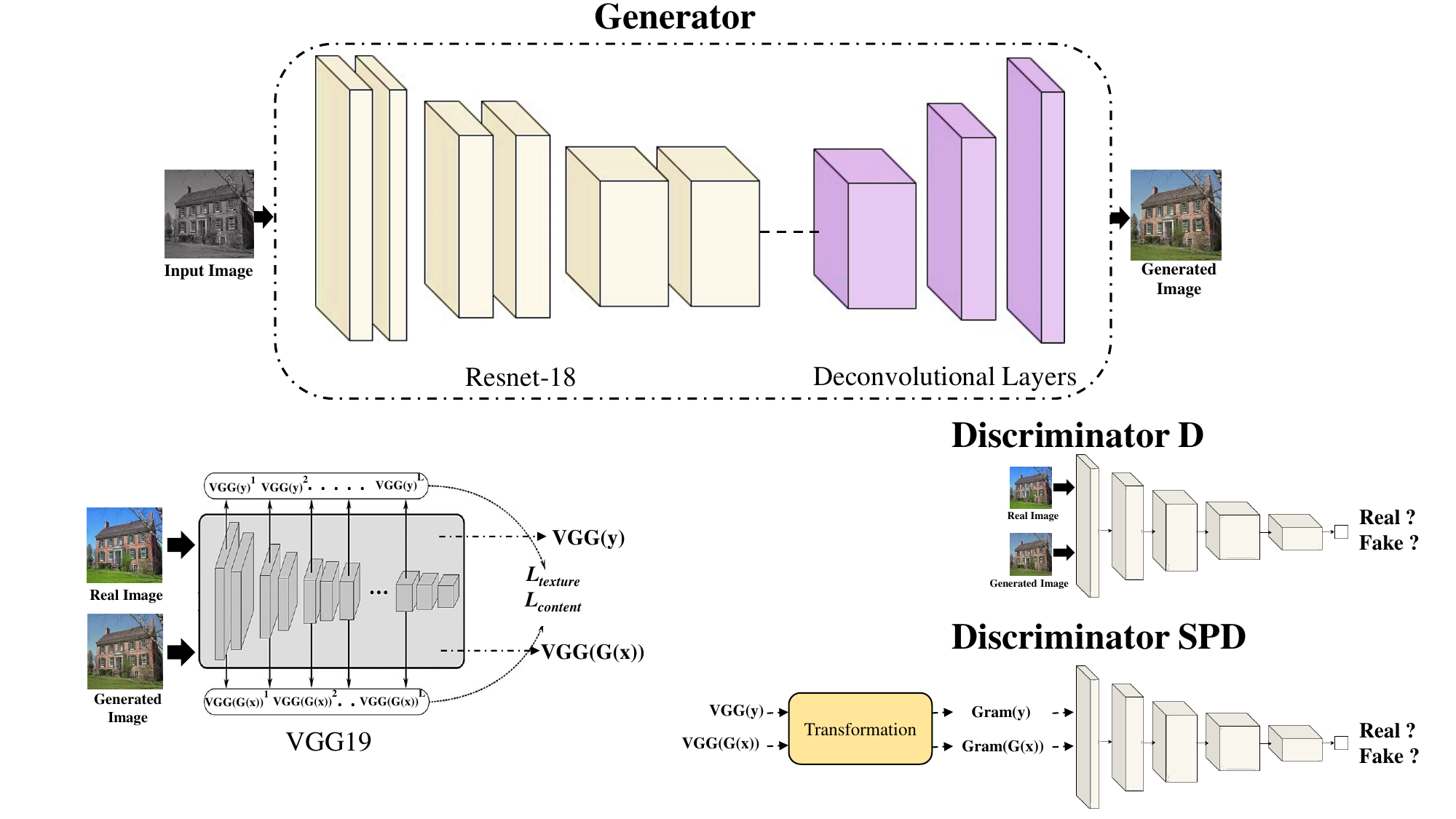}
\caption{Architecture of the proposed model. We use a feature discriminator on an SPD neural network and an image discriminator in the pixel domain. The terms $VGG(y)$ and $VGG(G(x))$ refer to the features of a pre-trained VGG-19 that were extracted from real images and generated images, respectively.} \label{fig:model}
\end{center}

\end{figure*}

The proposed model's design is presented in detail in this section. To generate a colorized image $I_g$ from a given grayscale image $I_{gray}$, with similar realistic colors to the original one (RGB), we propose a new GAN architecture that uses feature images in both the discriminator and the generator. The main ideas of our work are the following: first, our generator employs a ResNet architecture with a few alterations by adding a deconvolutional network; then, to discriminate between fake and real images we use two discriminators from different domains. The first one is for the image pixel domain, and the second one is for the feature image domain based on an SPD neural network. We propose adversarial losses during training to provide realistic colorized outputs. In the part that follows, we go into more depth about our improved network structure, whose main network structure is shown in Fig. \ref{fig:model}.

\subsection{Image Generator \& Discriminator Networks}

\begin{figure}[htbp]
\centering
\includegraphics[width=8cm,height=5cm]{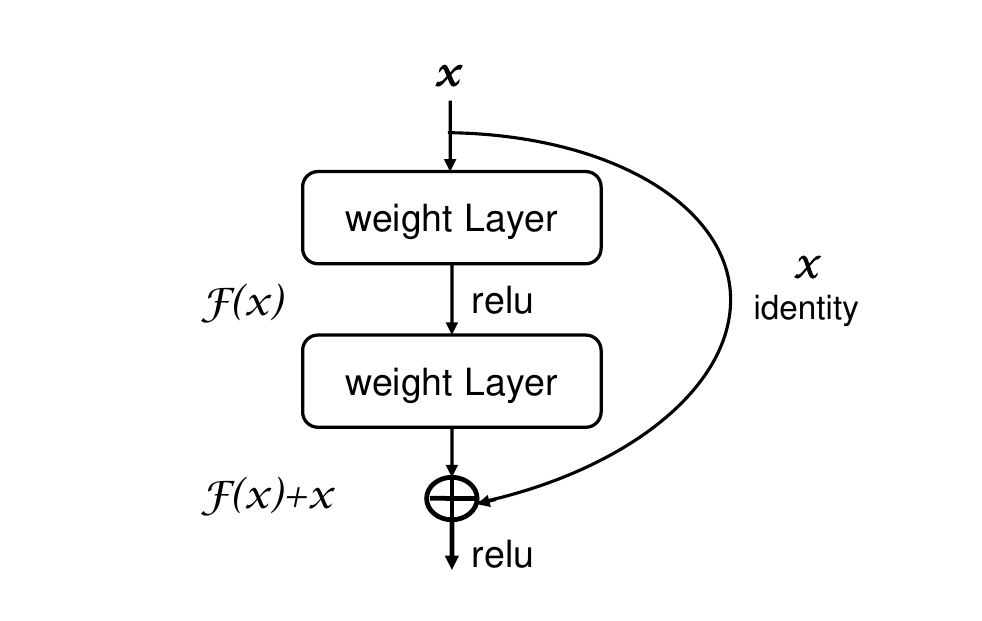}
\caption{Example of building blocks for ResNet architecture.} 
\label{fig:resnet}
\end{figure}

The problem with deep networks is the reduction of the gradient as the back-propagation progresses. Indeed, at each layer, the gradient will decrease slightly. While this is not a problem for networks with a few layers. For our generator, we choose ResNet architecture which consists of down-sampling, residual blocks, and up-sampling parts (Deconvolutionnal layers, Fig. \ref{fig:model}). We use this architecture because it employs residual connections that prevent the vanishing gradient across layers. ResNet adds a connection from the convolutional input directly to the output. This creates two “paths” for propagation and especially backpropagation. Fig. \ref{fig:resnet} shows ResNet’s basic building block. Then, a deconvolutional network is employed which takes as an input the output features generated by Resnet. Each layer in the deconvolutional network consists of a  transposed convolutional for upsampling. The last layer of the network is a $1 \times 1$ convolution followed by the tanh function. The number of channels in the output layer is 3 with L*a*b* color space of the generated image. While the generator network G tries to fool the discriminator network D by generated images, D takes the generated and real images and tries to distinguish between them. As long as D is able to effectively discriminate its input, G benefits from the gradient that the D network offers through its adversarial loss. Our discriminator network employs five convolutional layers followed by spectral normalization, as used in pix2pix, which is called  $70 \times 70$ PatchGAN. The reason why we use a PatchGAN discriminator, it is because it proved a better quality score than the CNN discriminator which do not use spectral normalization.

\begin{figure*}[h!]
\centering
\includegraphics[width=13cm,height=4cm]{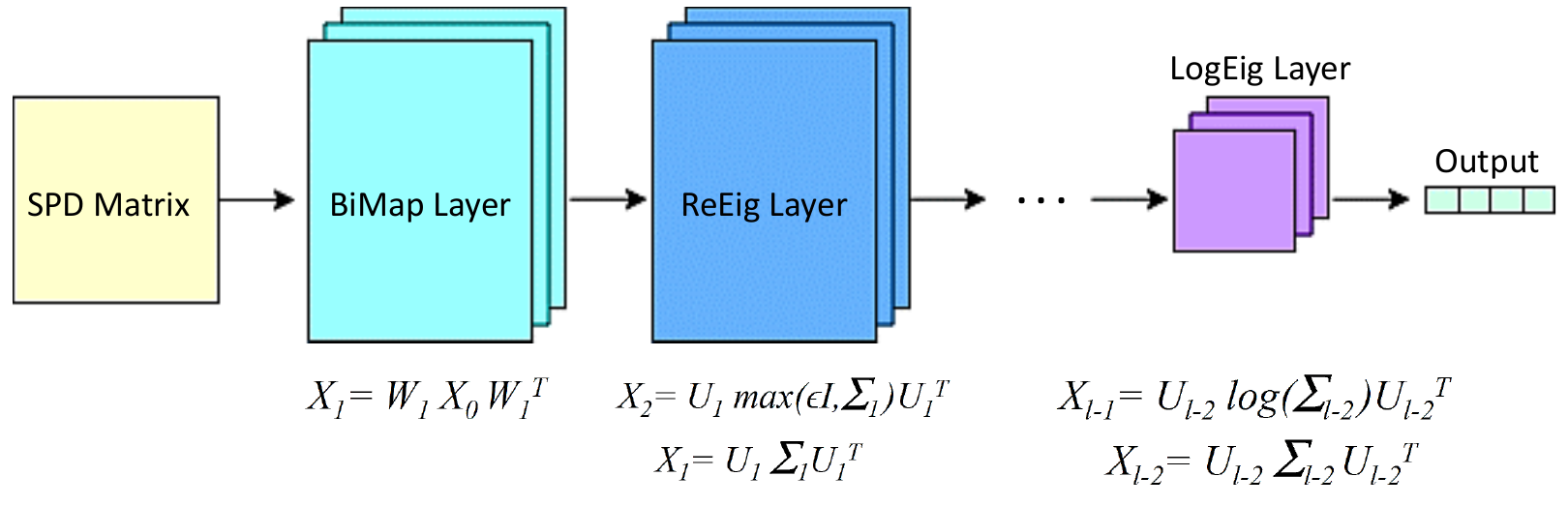}
\caption{Architecture of the proposed SPD discriminator network.} 
\label{fig:SPD}
\end{figure*}

\subsection{Riemannian SPD Network Discriminator}

As discussed previously, traditional GANs are based on CNN which consists of convolutional layers, max or average pooling that only capture the first-order information. For easier training, the ReLU function becomes the default activation function for many convolutional networks. Moreover, it introduces non-linearity but does so only at individual pixel level. SPD matrices computed from image features are believed to be better able to capture regional features than first-order statistics. Recent works \cite{li2017demystifying} show that style loss (based on SPD matrices) computes the difference between some aspects of images such as global color information. Furthermore, by plotting the image histograms (the total and each color channel separately) of the content reference, style reference, and generated images, the distribution of colors in the generated image matches more closely the style reference, as opposed to the content reference. This is the reason why we can employ the SPD matrices in the context of color transfer. Considering that color is highly correlated with image visual quality, a natural idea is to introduce a discriminator dedicated to SPD matrices that can be used in identification which can help to avoid color misalignment.

For all these reasons, we decide to employ the SPD network (SPDNet), which is composed of bilinear mapping (BiMap) and eigenvalue rectification (ReEig) layers that replace the fully connected convolution-like layers and rectified linear units (ReLU)-like layers. The underlying idea of using BiMap layers is to create new SPD matrices with a bilinear mapping from the input SPD matrices, which are typically covariance matrices produced from image characteristics. The generated SPD matrices are then rectified with a non-linear function by the ReEig layers. Due to the SPD matrices' non-Euclidean manifolds, an eigenvalue logarithm (LogEig) layer is used to perform Riemannian computing on them and produce their Euclidean forms for any regular output layers. Fig. \ref{fig:SPD} depicts the SPD network's architecture.

The BiMap layer, which creates new SPD matrices from the input SPD matrices using a bilinear mapping function $f_b$, is employed in the SPD network discriminator to build more compact and discriminative SPD matrices.

\begin{equation}
X_k=f_b^{(k)}(X_{k-1},W_k)=W_k X_{k-1} W_k^T,
\end{equation}

where $X_{k-1}$ is the input SPD matrix of the k-th layer, $W_k \in \mathbb R_*^{d_k x d_{k_1}}$. It should be noted that multiple bilinear mappings can be also performed on each input.

To improve discriminative effectiveness, rectified linear units (ReLU) have been used in convolutional neural networks (CNNs). The SPD network develops a non-linear function $f_r$ for the ReEig (k-th) layer in order to rectify the SPD matrices by tuning up their small positive eigenvalues, which is motivated by the notion of the non-linearity of ReLU:

\begin{equation}
X_k=f_r^{(k)}(X_{k-1})=U_{k-1} max(\epsilon I, \sum_{k-1}) U_{k-1}^T,
\end{equation}

where $U_{k-1}$ and $\sum_{k-1}$ are obtained using an eigenvalue decomposition, $X_{k-1}$ can be defined by:

\begin{equation*}
X_{k-1}=U_{k-1}\sum_{k-1}U_{k-1}^T,
\end{equation*}

$\epsilon$ denotes a rectification threshold, $I$ is an identity matrix, $max(\epsilon I, \sum_{k-1})$ refers to the diagonal matrix A with diagonal elements being defined as follows:

\[A(i,j)=
\left\{
\begin{array}{r c l}
\sum_{k-1}(i,i)&,& \sum_{k-1}(i,i) > \epsilon\\
\epsilon &,& otherwise
\end{array}
\right.
\]

For output layers with objective functions, it is suggested that the LogEig layer performs Riemannian computation on the generated SPD matrices. The authors of \cite{arsigny2007geometric} demonstrate that Log-Euclidean Riemannian metric can endow the Riemannian manifold of SPD matrices with a Lie group structure, in order to ensure that the manifold is reduced to a flat space when the matrix logarithm operation log(.) is applied on the SPD matrices. Therefore, the Riemannian computation \cite{arsigny2007geometric} was used to define the relevant function $f l$:

\begin{equation}
X_k=f_l^{(k)}(X_{k-1})=log(X_{k-1})=U_{k-1} log(\sum_{k-1}) U_{k-1}^T,
\end{equation}

where $X_{k-1}=U_{k-1}\sum_{k-1}U_{k-1}^T$ denotes an Eig operation, and $log(\sum_{k-1})$ refers to the diagonal matrix of eigenvalue logarithms.

Authors in \cite{huang2017riemannian} propose to use a traditional Euclidean network layer after applying the LogEig layer, such as the Euclidean fully connected (FC) layer which could be inserted after the LogEig layer. In our case, we will take only the output of the LogEign layer because it contains much more information about the transformation applied, and without using any Fully Connected (FC) layers, we keep the information that can be lost by nonlinear projection and losses.

To train the SPD network, we adopt the back-propagation method described in \cite{huang2017riemannian} which proposes a new way of updating the weights in different SPD network layers.

\subsection{Proposed Losses}

To train the architecture of GAN so that it can effectively achieve image colorization, the generator network is employed to colorize images from grayscale ones. The discriminator network determines if color image data is generated artificially or naturally. This can be expressed as follows:

\begin{equation}\label{eqgan1}
\begin{array}{ll}
L_{GAN}(G,D) &=\underset{y\sim pdata(y)}{\mathbb E}[\log(D(y)]\\
&+\underset{x\sim pdata(x)}{\mathbb E}[\log(1-D(G(x))]
\end{array}
\end{equation}

The GAN model tries to solve the minimax problem which is defined as follows:

\begin{equation}
\begin{array}{ll}
\underset{G}{\min}\textbf{ }\underset{D}{\max}\textbf{ }(&\mathbb E_{y \sim p_{data}(y)}[\log(D(y))]+\\& \mathbb E_{x \sim p_{x}(x)} [\log(1-D(G(x)))])
\end{array}
\end{equation}

Where y is the color image from a real distribution, x is the grayscale image, G(x) is the output of a generator network, and D is the discriminator network. We further refer to data distributions as $y\sim pdata(y)$ and $x\sim pdata(x)$. \\

\textbf{$L_{l1}$ Loss:}

In the context of automatic colorization, to allow the network to predict the perceptually plausible colors, we adopt the $l_1$-loss. Given two images $I_{generated}$ and $I_{real}$, where $I_{generated}$ is the generated image and $I_{real}$ is the original one, the per-pixel $l_1$-loss can be written as:

\begin{equation}\label{l1}
L_{l1}= \underset{x,y}{\mathbb E}||y-G(x)||_1
\end{equation}

where $G(x)$ and $y$ have the same size. \\

\textbf{Multi-discriminator Loss $L_{M-Dis}$}:

Based on the above loss functions, the generator is trained with the two discriminators using a loss function that takes the following form : 

\begin{equation}\label{M-Dis}
\begin{array}{ll}
L_{M-Dis}(G,D_i,D_{SPD}) &=\lambda_{i}L_{GAN}(G,D_i)+ \\& \lambda_{SPD}L_{GAN}(G,D_{SPD})
\end{array}
\end{equation}

where $\lambda_{i}$ and $\lambda_{SPD}$ refer to defined regularization factors, $L_{GAN}(G,D_i)$ denotes a pixel GAN loss that refers to high-frequency details in the pixel domain, $L_{GAN}(G,D_{SPD})$ represent the color GAN loss that characterizes color details.\\

The pixel GAN loss can be formulated by the equation:
\begin{equation}
\begin{array}{ll}
L_{GAN}(G,D_i) &=\underset{y\sim pdata(y)}{\mathbb E}[\log(D_i(y)]\\
&+\underset{x\sim pdata(x)}{\mathbb E}[\log(1-D_i(G(x))]
\end{array}
\end{equation}

To compute the SPDGAN loss, for the discriminator $D_{SPD}$, we take as input of the discriminator the gram matrix to discriminate images by the distribution of colors. The gram matrix of the image feature Gram(x) extracted from the VGG19, can take the formula below:

\begin{equation}\label{gram}
Gram(x) =\sum_{i,j=1}^{m} F_i(x) F_j(x)
\end{equation}

where $F_i$ and $F_j$ indicate respectively the ith and jth feature map in the extracted multi-resolution feature maps. In this work, the feature map has been flattened to the matrix. Hence, the gram matrix  can be calculated as follow:

\begin{equation}\label{gram}
Gram(x)= F(x) F(x)^{T}
\end{equation}

So, our SPDGAN loss can be defined by:

\begin{equation}
\begin{array}{ll}
L_{GAN}(G,D_{SPD}) &=\underset{y\sim pdata(y)}{\mathbb E}[\log(D_{SPD}(Gram(y)]\\
&+\underset{x\sim pdata(x)}{\mathbb E}[\log(1-D_{SPD}(Gram(G(x)))]
\end{array}
\end{equation}

\begin{figure}[!h]
\centering
\includegraphics[width=8cm,height=5cm]{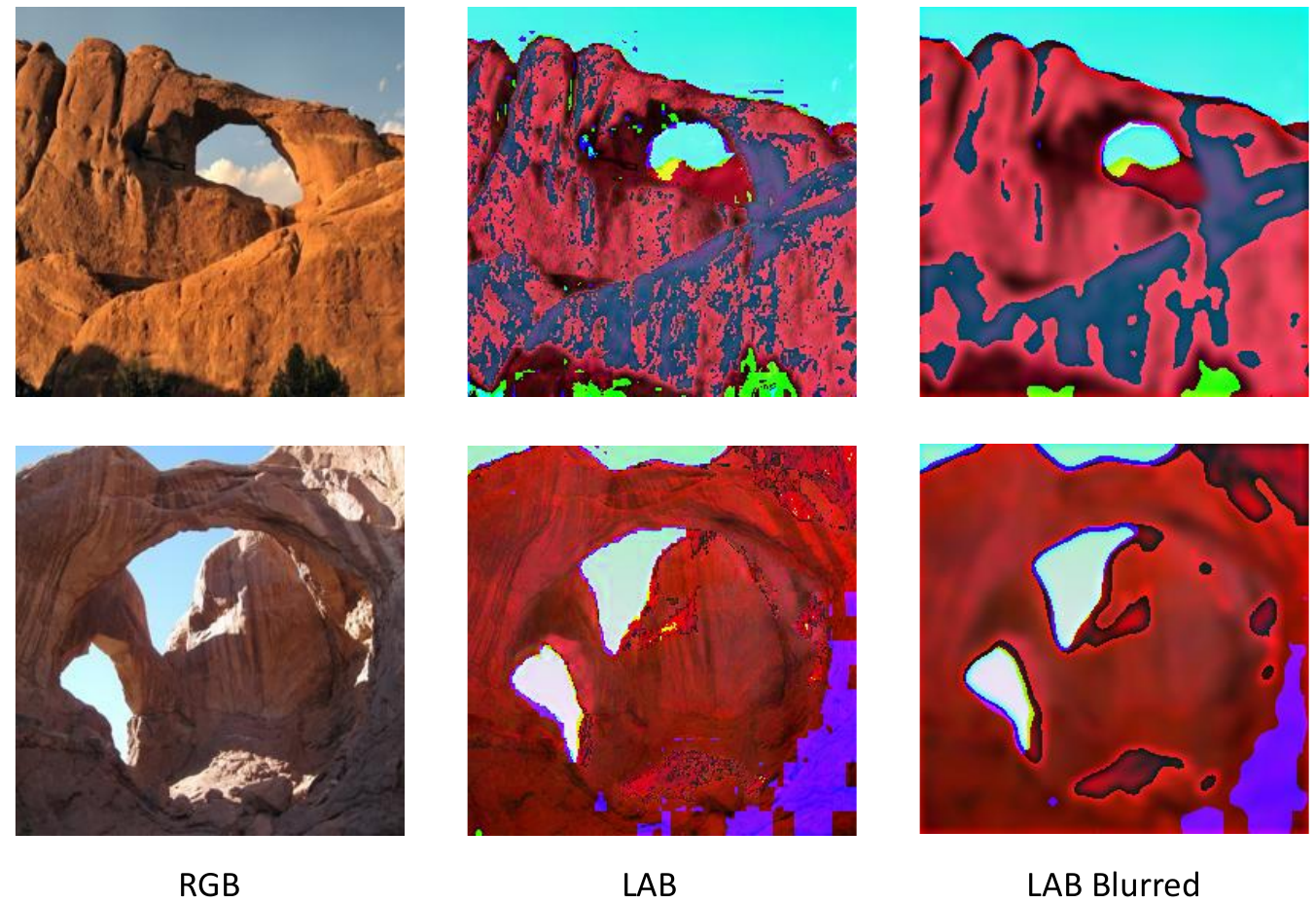}
\caption{Results of an RGB image after applying the Gaussian filter $B$, on its LAB transformation with  $\sigma_{x,y}=3$.}
\label{fig:blurlab}
\end{figure}

\textbf{Color loss $L_{color}$}: 

In addition, to compare the difference in brightness, contrast, and major colors between images, we suggest non-adversarial color loss. It can be written as follows:

\begin{equation}\label{color}
L_{color} =\lVert y_b-G(x)_b\rVert_2^2
\end{equation}

where $y_b$ and $G(x)_b$ are the blurred images of $y_{LAB}$ and $G(x)_{LAB}$, resp.\\

The blurred image is obtained via blurred convolution $B$:

\begin{equation*}
y_b= y_{LAB} * B, \textbf{  } G(x)_b= G(x)_{LAB} * B
\end{equation*}

where $B$  refers to the blur filter and * stands for the convolution process. The size of the filter $B$ is $21 \times 21$ with stride 1, and the weights of the filter $B$ are fitted to the Gaussian distribution. The following equation can be used to calculate $B$:

\begin{equation}\label{eq8}
B(x,y) = A \exp (-\frac{(x-\mu_x)^2}{2\sigma_x^2} -\frac{(y-\mu_y)^2}{2\sigma_y^2}
\end{equation}

with $A=0.053$, $\mu_{x,y}=0$ and $\sigma_{x,y}=3$. By visual inspection, we settled on the constant $\sigma_{x,y}$ as the minimum value that guarantees that texture and content are dropped as seen in Fig. \ref{fig:blurlab}. We choose the LAB color space instead of RGB because the distances between colors in LAB space match the perceptual distances of colors for a human observer.

\textbf{Full Objective}: 

The following equation can be used to express our full objective loss:

\begin{equation}\label{Full}
\begin{array}{ll}
L_{Multi-GAN}(G,D_i,D_{SPD})&=L_{M-Dis}(G,D_i,D_{SPD})\\ &+ \lambda_{l1} L_{l1} + \lambda_{color}L_{color}
\end{array}
\end{equation}

where $\lambda_{l1}$ and $\lambda_{color}$ are parametric factors. Our objective is to use the value function to resolve the minimax game problem:

\begin{equation}
\begin{array}{ll}
G,D_i,D_{SPD} = &arg \textbf{ } \underset{G}{\min} \underset{D_i,D_{SPD}}{\max} \\& L_{Multi-GAN}(G,D_i,D_{SPD}) 
\end{array}
\end{equation}

\subsection{Normalization techniques}

GANs are highly effective deep learning architectures that can produce very sharp generated data, even for images with complex, highly multimodal distributions. Nevertheless, GANs are recognized to have major challenges in the training, and often suffer from some limitations such as mode collapse, non-convergence, and instability. Several normalization techniques have been proposed to address these training problems such as Batch Normalization, Spectral Normalization, and Instance Normalization.

\subsubsection{Batch Normalization}

Ioffe \emph{et al.} \cite{ioffe2015batch} proposed a Batch Normalization (BN) method to accelerate the training of deep networks. BN normalizes the features extracted across layers to have zero mean and unit variance to avoid the mode collapse. The latter refers to the mode where the generator creates samples with the same looking and very low diversity for different input images.

By taking a batch of samples $\{x_1,x_2,…,x_m\}$ we compute:

\begin{align*} 
    \mu_{\beta} \leftarrow \frac{1}{m} \sum_{i=1}^{m} x_i &\textit{   } \textit{   }  \textit{           // mini-batch mean}\\
    \sigma_{\beta}^2 \leftarrow \sum_{i=1}^{m} (x_i-\mu_{\beta})^2 &\textit{   } \textit{   }  \textit{           // mini-batch variance}\\
    \hat{x_i}  \leftarrow \frac{x_i-\mu_{\beta}}{\sqrt{\sigma_{\beta}^2+\epsilon}}  &\textit{   } \textit{   }  \textit{           // normalize} \\
    y_i \leftarrow \gamma \hat{x_i} +\beta  \equiv BN_{\gamma,\beta}(x_i) &\textit{   } \textit{   }  \textit{           // scale and shift}
\end{align*}

$\mu_{\beta}$ and $\sigma_{\beta}$ refer to the means and standard deviations of the input batch. \\
$\gamma$ and $\beta$ denote the learned parameters. After applying BN, the output will always have a mean $\beta$ and a standard deviation $\gamma$, regardless of the input distribution.

\subsubsection{Spectral Normalization}

it is a normalization approach introduced by Miyato \emph{et al.} \cite{miyato2018spectral} used in the context of GANs for training stabilization of the discriminator. 

It adjusts the Lipshitz constant of the discriminator $f$ by constraining the spectral norm of each layer $g:h_{in} \rightarrow h_{out}$. The useful aspect of this method is that the only hyper-parameter that needs to be modified is the Lipschitz constant. The hyper-parameter $||g||_{Lip}$  has the value $sup_h \sigma(\nabla g(h))$, where $sigma(a)$ denotes the spectral norm of the matrix $A$($L_2$ matrix norm of $A$):

\begin{equation*}
    \sigma(a) = \underset{h:h\neq0}{max} \frac{||Ah||_2}{||h||_2}=\underset{||h||_2\leq1}{max}||Ah||_2
\end{equation*}

which is equivalent to the largest singular value of $A$. Therefore for a linear layer $g(h)=Wh$ the norm is given by $||g||_{Lip} = sup_h \sigma(\nabla g(h)) = sup_h \sigma(W) = \sigma(W)$. Spectral normalization normalizes the spectral norm of the weight matrix $W$ so it satisfies the Lipshitz constraint $\sigma(W) = 1$ :

\begin{equation*}
    W_{SN}(W) = W/\sigma(W)
\end{equation*}

\subsubsection{Instance Normalization}

This technique was proposed by Ulyanov \emph{et al.} \cite{ulyanov2016instance}. it is also known as contrast normalization layer where:

\begin{align*} 
    y_{tijk}=&\frac{x_{tijk}- \mu_{ti}}{\sqrt{\sigma_{ti}^2+\epsilon}} \textit{  ,  } \\
    \mu_{tijk}=& \frac{1}{HW}\sum_{l=1}^W \sum_{m=1}^H x_{tilm} \textit{  ,  } \\  
    \sigma_{ti}^2 =& \frac{1}{HW} \sum_{l=1}^W \sum_{m=1}^H (x_{tilm}-mu_{ti})^2
\end{align*} 

This approach allows us to remove instance-specific contrast information from the content image, which simplifies generation and the learning process.

We describe in Section \ref{sec:exp} the effect of each normalization technique used on the generator and discriminator. Our goal is to select the adequate normalization strategy for each architecture (Generator and Discriminator) with selective parameters that conduct the best performance.

\subsection{SPDGAN network architecture}

In this section, we present the detailed layer description of the generator and the two discriminators. 

To overcome the limitation of GAN traditional architecture, we propose a residual-based network based on the Resnet architecture for the generator. The network constitutes 2 encoding blocks with 9 residual blocks and 2 decoding blocks. Each of the blocks (Encoding/Decoding) follows the two-stride Convolution-or-Deconvolution$\rightarrow$Batch-or-Instance Normalization$\rightarrow$Relu scheme.

For the image discriminator, we use PatchGAN architecture based on local patches of size 70$\times$70 instead of the hole image. The role of this discriminator is to distinguish whether an image is real or fake. It takes in two images, the grayscale image and (generated image or ground truth image), pass them across 5 down-sampling convolutional$\rightarrow$Batch-or-Spectral-or-Instance Normalization$\rightarrow$LeakyReLU layers, and outputs a matrix of size 30$\times$30. Each element in this matrix denotes the classification of one patch. 

For the SPD discriminator, we adopt the SPD network which 
takes the gram matrix constructed from extracted features of the VGG19
for grayscale images, generated images, and ground truth images. The network is made up of three blocks 1-Bloc$\rightarrow$2-Bloc$\rightarrow$3-Bloc$\rightarrow$LogEig, each bloc corresponding to BiMap$\rightarrow$ReEig operations.

\section{Experimental Analysis}
\label{sec:exp}

To evaluate its performance, our SPDGAN model has been extensively tested in comparison to existing state-of-the-art techniques. We start by the description of datasets and metrics used to evaluate our model. Then, we compare the performance obtained by different combination strategies in our proposed architecture. We also investigate the effectiveness of each component's contribution to our network through an ablation study. All these previous studies will be performed on Places365 \cite{zhou2017places} Dataset. Finally, we qualitatively compare our SPDGAN with several state-of-the-art methods on publicly available Places365 \cite{zhou2017places} and COCO-Stuff \cite{caesar2018coco} datasets. For quantitative comparisons, three metrics are used to evaluate the colorization results.

\subsection{Dataset and metrics}

\textbf{Dataset:} 
We conduct our experiments on two publicly available datasets:
\begin{itemize}
    \item \textit{Places365 \cite{zhou2017places} }:  a multiclass dataset that contains 365 categories of photos including natural scenery, humans, buildings, and so on. We divide the dataset into three sets: training set, validation set, and test set. There are 1.8 million images in the training set, 18 thousand images in the validation set, and 320 thousand images in the test set. All the images have the same resolution 256 $\times$ 256.
    
    \item \textit{COCO-Stuff \cite{caesar2018coco}}: is a subset of the COCO dataset that contains a wide variety of natural scenes with multiple objects present in the image, with more than 118K images for training and 5K images for the validation set. We use the 5,000 images in the original validation set for evaluation. All the images are resized to 256 $\times$ 256.

\end{itemize}
We train separate networks for Places365 \cite{zhou2017places} and COCO-Stuff \cite{caesar2018coco}for all the models (proposed and state-of-the-art). To evaluate the performance of models, we built 1000 images from the validation sets of the two datasets.

\textbf{Metrics:} we use a variety of comprehensive evaluation indicators to evaluate the colorization results. The selected indicators are Peak Signal-to-Noise Ratio (PSNR), Structural Similarity (SSIM), Frechet Inception Distance (FID), and Colorfulness Score. \\

\begin{itemize}






\item Frechet Inception Distance (FID) \cite{heusel2017gans} is a metric used to compute the distance between feature vectors calculated for real and generated images. This distance refers to how similar the two images are in terms of statistics on computer vision features by employing the inception v3 model used for image classification. Generally, FID is used to evaluate the quality of images generated by GANs, and lower scores have been shown to correlate well with higher-quality images.

The FID metric is computed by the following formula :

\begin{equation}
    d^2 = ||\mu_1 – \mu_2||^2 + Tr(C_1 + C_2 – 2*\sqrt(C_1*C_2))
\end{equation}

The score is referred to as $d^2$, showing that it is a distance and has squared units.
The $mu_1$ and $mu_2$ refer to the feature-wise mean of the real and generated images. The $C_1$ and $C_2$ are the covariance matrix for the real and generated feature vectors.

\item Colorfulness Score \cite{hasler2003measuring} represents the vividness of generated images. We start computing the Colorfulness score using the equations below:

\begin{equation*}
\begin{array}{ll}
    rg =& R - G \\
    yb = &\frac{1}{2}(R + G)-B
\end{array}
\end{equation*}

where R, G, and B denote Red, Green, and Blue color space respectively. Next, the standard deviation ($\sigma_{rgyb}$) and mean ($\mu_{rgyb}$) are computed before calculating the final colorfulness metric, C.

\begin{equation*}
\begin{array}{ll}
    \sigma_{rgyb} =&\sqrt{\sigma_{rg}^2+\sigma_{yb}^2} \\
    \mu_{rgyb} =&\sqrt{\mu_{rg}^2+\mu_{yb}^2}\\
    C =& \sigma_{rgyb} + 0.3*\mu_{rgyb}
\end{array}
\end{equation*}

As we will find out, this turns out to be an extremely efficient and practical way of computing image colorfulness.
\end{itemize}

\subsection{Environmental Setup and Training details}

\textbf{Environment}: our model was developed on python 3.6 using the Pytorch framework. We used a PC with Intel® Xeon® Silver 4215R CPU, with 32GB of RAM and GeForce GTX 3080 Ti 16GB RAM graphics card. All the models are run on the same machine.\\

\textbf{Training details}: Adam is chosen as our generator and discriminator optimizer for our SPDGAN training. We conduct extensive experiments to select the adequate value of the initial learning for the  generator and the two discriminators (Image \& SPD) that leads  to the best colorization. Results show that with  $\alpha_G = 0.0003$ and  $\alpha_{D_{image}} = 0.00003$ and  $\alpha_{D_{SPD}} = 0.01$ we obtain a plausible colorization. We set the weight factors after extensive hyper-parameter optimization results of extensive hyper-parameters optimization. As result, we have $\lambda_i = 0.01$ and  $\lambda_{SPD} = 0.01$ in  the Equation \ref{M-Dis} and  $\lambda_{l_1} = 0.99$, $\lambda_{color} = 0.001$ in the Equation \ref{Full}.500 training epochs are considered, with a batch size of 4.

\begin{table*}[ht]
\begin{center}
\caption{Result of different combinations of normalization strategies between Generator and Discriminator .}\label{tab:normalization}  
\begin{tabular}{llllllllllll}
    \hline\noalign{\smallskip}  
       Generator  &  Discriminator & PSNR & SSIM &  FID      \\ 
            \noalign{\smallskip}\hline\noalign{\smallskip}           
          Batch Normalization  & Batch Normalization & 24.02 &0.92  & 3.97      \\
          & Instance Normalization&24.33  &0.91  & 4.02    \\
          & Spectral Normalization & 25.15 & 0.92 &  3.89      \\

    \hline  
         Instance Normalization  & Batch Normalization &23.89  &0.89  &  5.13      \\
          & Instance Normalization  &23.95  & 0.88 & 5.07       \\
          & Spectral Normalization & \textbf{26.12} & \textbf{ 0.95} &   \textbf{3.11}      \\

    \hline      
 \end{tabular} 
 \end{center}
 \end{table*}

\subsection{Effect of normalization techniques}

As mentioned previously, to avoid generated colorized images that suffer from visual artifacts we employ normalization techniques that help in the training process by reducing the Internal Covariate Shift (change in the distribution of network activation's due to the change in network parameters), and also for making the optimization faster. To demonstrate the importance of this step, we conduct several experiments by testing different combinations of normalization strategies between the generator and the discriminator architecture. We note that for the generator we only use the two normalization techniques: Batch and Instance Normalization without using Spectral Normalization which is only dedicated to discriminator architecture. We show in Table \ref{tab:normalization} the performance of each combination on the two databases. We conclude that the best combination of normalization strategy is instance normalization for the generator and spectral normalization for the discriminator, due to the observed high scores
of the different metrics PSNR, SSIM and FID. We justify this result by the fact that instance normalization performs style normalization by normalizing feature statistics, which have been found to carry the style information (Color) of an image, and by using the spectral normalization we stabilize the training of the discriminator. In the following, we adopt this combination for comparison with the state-of-the-art approaches.

\subsection{Effect of VGG-19 features}
As mentioned previously, we adopted in this work a VGG19 network to extract features from generated images and real ones. VGG19 is a very deep convolution neural network trained with ImageNet dataset and owns good versatility for features extraction and many works adopt it as the pre-trained model. We conduct several experiments to study the effect of extracted features by the VGG19 in the colorization task, and to select next the suitable layers for feature extraction. We adopt three features from $conv1-1$, $conv2-1$, and $conv3-1$.

\begin{figure*}[h!]
\centering
\includegraphics[width=13cm,height=7.5cm]{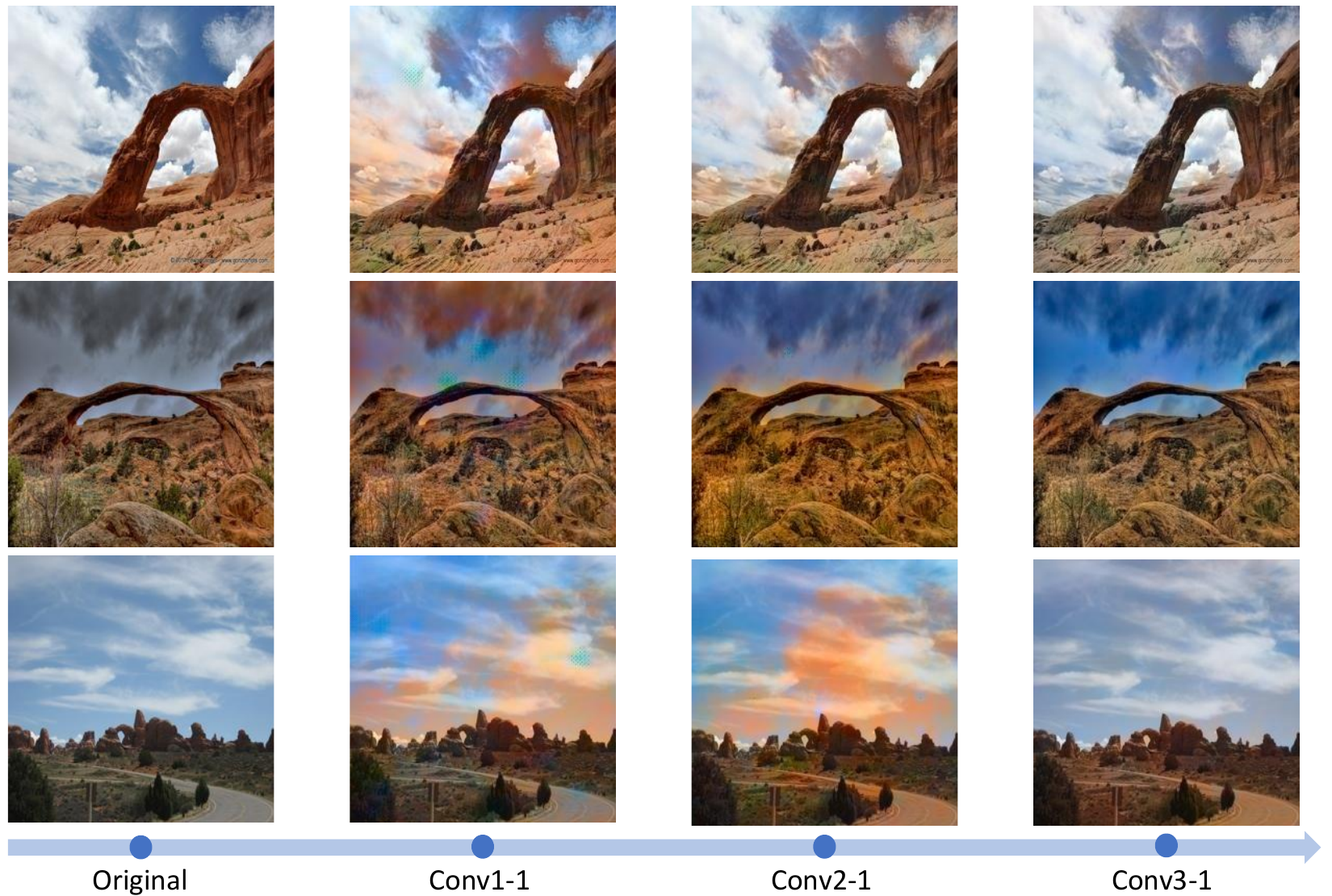}
\caption{Influence of the features extracted from different levels (conv1-1, conv2-1, conv3-1) of the VGG19 network in our colorization process.}
\label{fig:vgg19}
\end{figure*}

As we can clearly see from Fig. \ref{fig:vgg19}, the visually most appealing images are usually created by features extracted from high layers in the network, which is why we choose the features extracted from the conv3-1 layers that produce better and more reasonable colors results.

\subsection{Effect of blocs number on the SPD discriminator network}

\begin{figure}[h!]
\centering
\includegraphics[width=8cm,height=6cm]{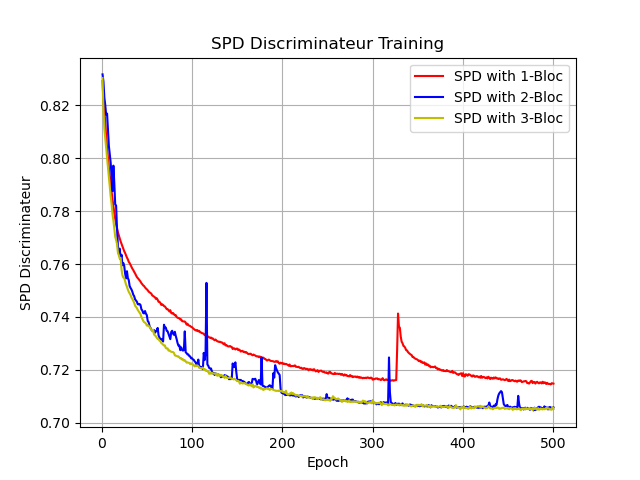}
\caption{Influence of the SPD discriminator blocs 1-Bloc / 2-Bloc / 3-Bloc in the training process.}
\label{fig:spdblocs}
\end{figure}

The main theoretical advantage of the proposed SPD discriminator is its ability for non-linear and deep learning mechanisms. In this work, our discriminator works with SPD matrices (e.g. Gram Matrices) which can be computed from the extracted features of the VGG19 network. In order to choose the best configuration of our SPD discriminator network, we study its behaviors with 3 different settings: 1-Bloc$\rightarrow$2-Bloc$\rightarrow$3-Bloc where each Bloc corresponds to BiMap$\rightarrow$ReEig operations, then, at the final step we apply the LogEig operation. We train our SPD discriminator with different configurations for 500 epochs. As shown in Fig. \ref{fig:spdblocs}, we report the performances of each configuration. Results show that our proposed SPDNet-3Bloc achieves several improvements and stabilize the training of the SPD discriminator network in term of $L_{GAN}(G,D_{SPD}) $ compared to the other configurations.

\subsection{Ablation Study}

In this section, we conduct an ablation study on Places365 datasets to assess the effectiveness of each component's contribution of our network. The settings of the ablation experiments are as follows:\\

(a)	Baseline: only the pixel discriminator is employed, which is based on a standard CNN architecture to distinguish if the colorized images are reals or fakes, while the Resnet architecture is used for the generator.\\

(b)	Baseline + SPD Discriminator: the SPD discriminator is added to the baseline architecture, which is used as a feature discriminator of the SPD matrices to avoid color misalignment in generated images. \\

(c) Baseline + SPD Discriminator + $L_{color}$ (SPDGAN): in this ablation experiment the color loss $L_{color}$ is considered. As mentioned previously this loss computes  the Euclidean distance between the blurred images from the generator and the ground truth. The idea behind using $L_{color}$ is to evaluate the difference in brightness, contrast, and major colors between the images
while eliminating the comparison of texture and content. Hence, we fixed a constant $\sigma$ by visual inspection as the smallest value that ensures that texture and content are dropped and forces the generated image to have the same color distribution as the ground truth one, while being tolerant to small mismatches.\\

\begin{figure*}[h!]
\centering
\includegraphics[width=13cm,height=4.5cm]{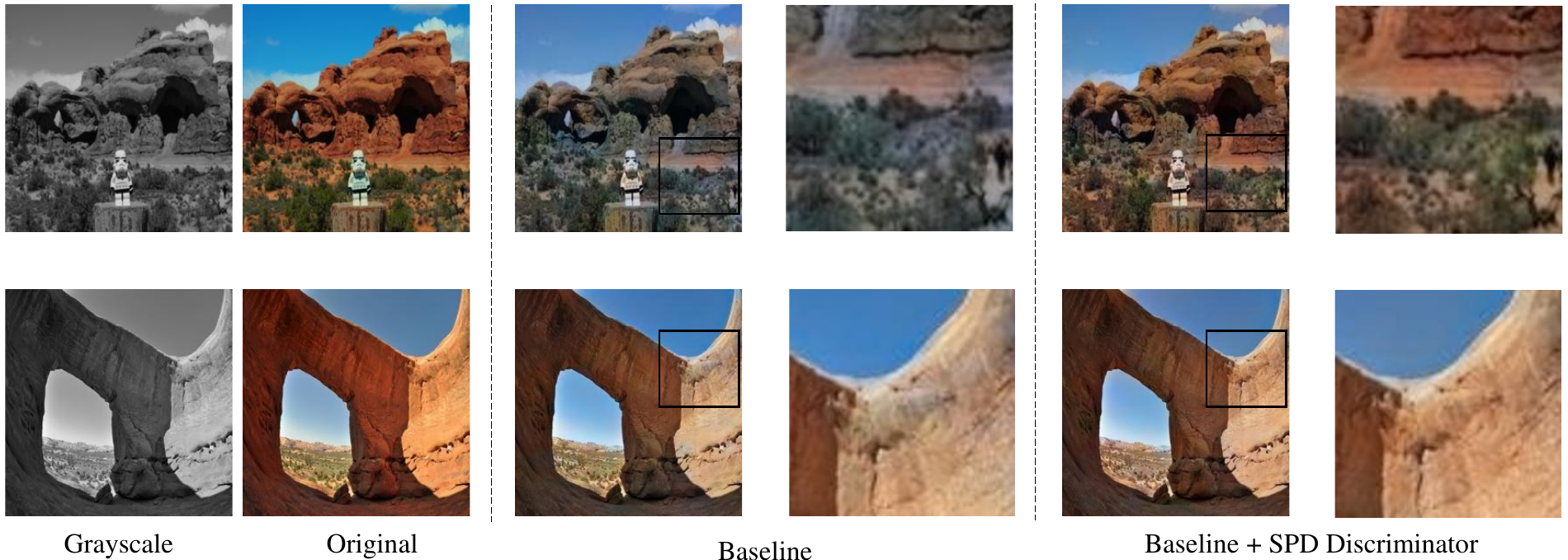}
\caption{Figure is better seen zoomed on the digital version of this document. Influence of the SPD Discriminator in our proposed model.}
\label{fig:abla1}
\end{figure*}

Some comparison results are presented in Fig. \ref{fig:abla1}. We can observe that our model with the SPD discriminator $D_{SPD}$ produces better and more stable colorization results compared to the case where we only use the baseline model, which generates colorized images with mismatched color results.

\begin{figure*}[h!]
\centering
\includegraphics[width=13cm,height=3cm]{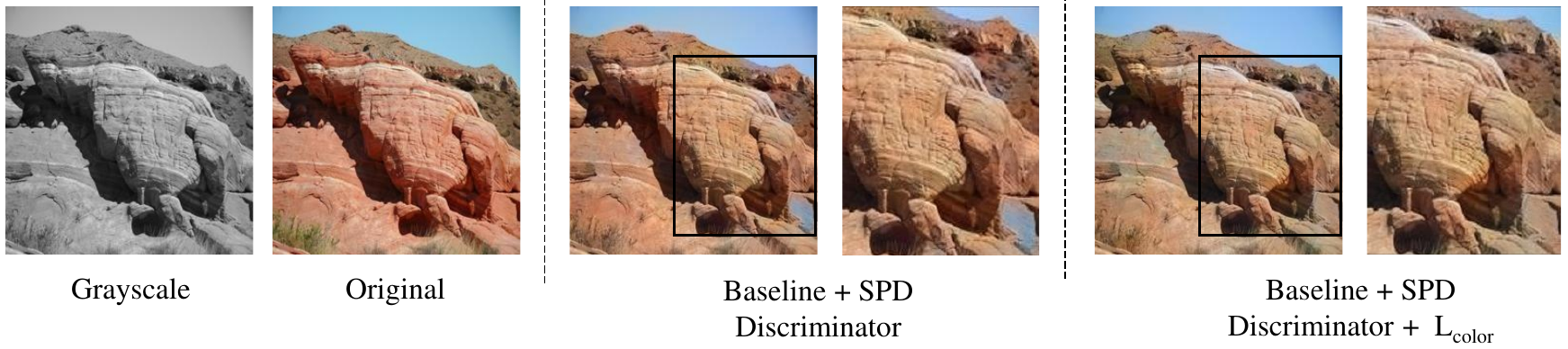}
\caption{Figure is better seen zoomed on the digital version of this document. Influence of $L_{color}$ loss in our proposed model.}
\label{fig:abla2}
\end{figure*}

Moreover, by adding the color loss, we improve the colorization result by producing more stable results as shown in Fig. \ref{fig:abla2} (Baseline + SPD Discriminator + $L_{color}$) without the small artifacts produced by the proposed model with the SPD Discriminator. As a result, we can observe that the color loss forces the colorized image to have the same color distribution as the ground truth one.

\subsection{Comparison with state-of-the-art methods}

In this section, the colorization results of our SPDGAN are evaluated by visual inspection and quantitative metrics against several state-of-the-art methods. Finally, a subjective user study is also performed to quantitatively evaluate the results of different methods.

\subsubsection{Quantitative comparisons and visual inspection}

Finally, with the appropriate parameters and configurations discussed previously, we perform a quantitative and qualitative comparison of our proposed SPDGAN with state-of-the-art methods on the Places365 and COCO-Stuff datasets.

Our SPDGAN was compared to four robust colorization approaches Pix2Pix \cite{isola2017image}, Nazeri \emph{et al.} \cite{nazeri2018image}, DeOldify \cite{DeOldify} and  ChromaGAN \cite{vitoria2020chromagan}. The quantitative results on four metrics are reported in Tables \ref{tableplaces365} and \ref{tablecoco}. 

For the colorfulness score, our model obtain the highest colorfulness score as observed in Tables \ref{tableplaces365} and \ref{tablecoco} compared to comparison methods. In terms of FID SODGAN achieves the lowest FID, which means that it could generate colorization results with better image quality and fidelity. We also evaluate the PSNR and SSIM values between colorized images and ground truth, even if it is well-known
that such pixel-wise measurements may not well reflect the colorization performance, as plausible colorization results probably diverge a lot from the original color image. The results are reported in Tables \ref{tableplaces365} and \ref{tablecoco}, which show that we outperform largely the other methods.

\begin{table}[h!]
\begin{center}
\caption{Quantitative comparison on COCO-Stuff \cite{caesar2018coco} dataset}
\begin{tabular}{llllllllllll}
    \hline\noalign{\smallskip}  
         & PSNR$\uparrow$ & SSIM$\uparrow$ &  FID$\downarrow$ & Colorful$\uparrow$     \\ 
            \noalign{\smallskip}\hline\noalign{\smallskip}           
          DeOldify \cite{DeOldify}  & 21.985 & 0.919 & 3.952 & 28.213      \\
          ChromaGAN \cite{vitoria2020chromagan} & 21.521 & 0.890 & 5.097 & 23.415   \\
          Pix2Pix \cite{isola2017image} & 20.428 & 0.794 & 6.125& 21.120    \\

         Nazeri \emph{et al.} \cite{nazeri2018image} & 24.706 & 0.938 & 3.571 &  34.384     \\
         SPDGAN (ours) & \textbf{26.174} &\textbf{0.964} & \textbf{3.097} &   \textbf{37.997}     \\
        
    \hline      
 \end{tabular} 
 \label{tableplaces365}
 \end{center}
 \end{table}

\begin{table}[h!]
\begin{center}
\caption{Quantitative comparison on Places365 \cite{zhou2017places} dataset} 
\begin{tabular}{llllllllllll}
    \hline\noalign{\smallskip}  
       & PSNR$\uparrow$ & SSIM$\uparrow$ &  FID$\downarrow$ & Colorful$\uparrow$     \\ 
            \noalign{\smallskip}\hline\noalign{\smallskip}           
          DeOldify \cite{DeOldify}  & 21.933 & 0.939 & 3.922 & 28.101      \\
          ChromaGAN \cite{vitoria2020chromagan} & 23.479 & 0.873 & 5.077 & 23.431   \\
          Pix2Pix \cite{isola2017image} & 20.356 & 0.729 & 5.978& 21.011    \\

         Nazeri \emph{et al.} \cite{nazeri2018image} & 24.637 & 0.921 & 3.583 &  34.357     \\
         SPDGAN (ours) & \textbf{26.125} &\textbf{0.957} & \textbf{3.115} &   \textbf{37.962}     \\
        
    \hline      
 \end{tabular} 
 \label{tablecoco}
 \end{center}
 \end{table}

\begin{figure*}[h!]
\centering
\includegraphics[width=13cm,height=17cm]{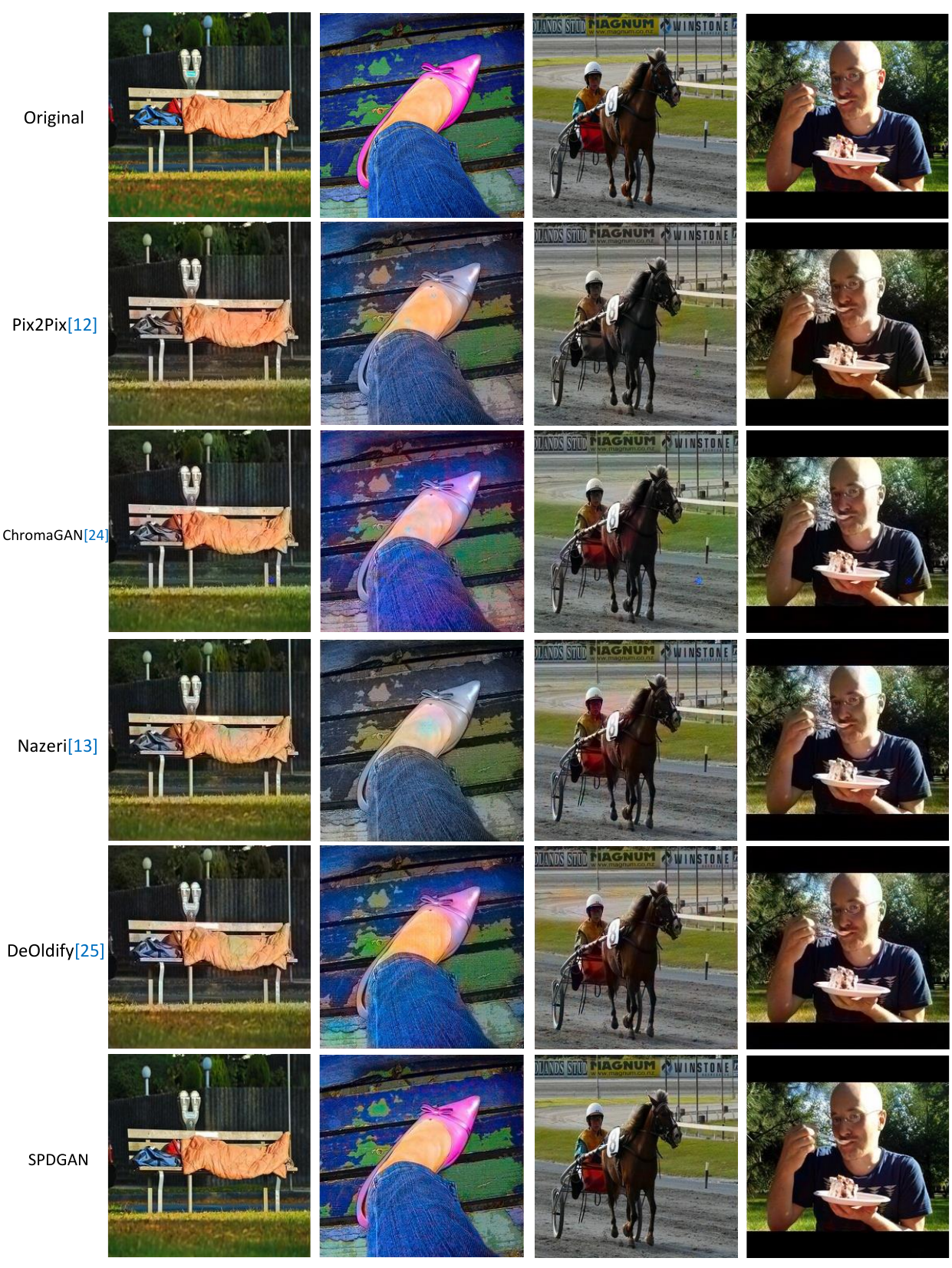}
\caption{Figure is better seen zoomed on the digital version of this document.  Qualitative comparisons of our proposed SPDGAN with state-of-the-art approaches on COCO-Stuff \cite{caesar2018coco} dataset. Our SPDGAN is able to generate more realistic images with plausible color results.}
\label{fig:coco}
\end{figure*}

\begin{figure*}[!ht]
\centering
\includegraphics[width=13cm,height=16.5cm]{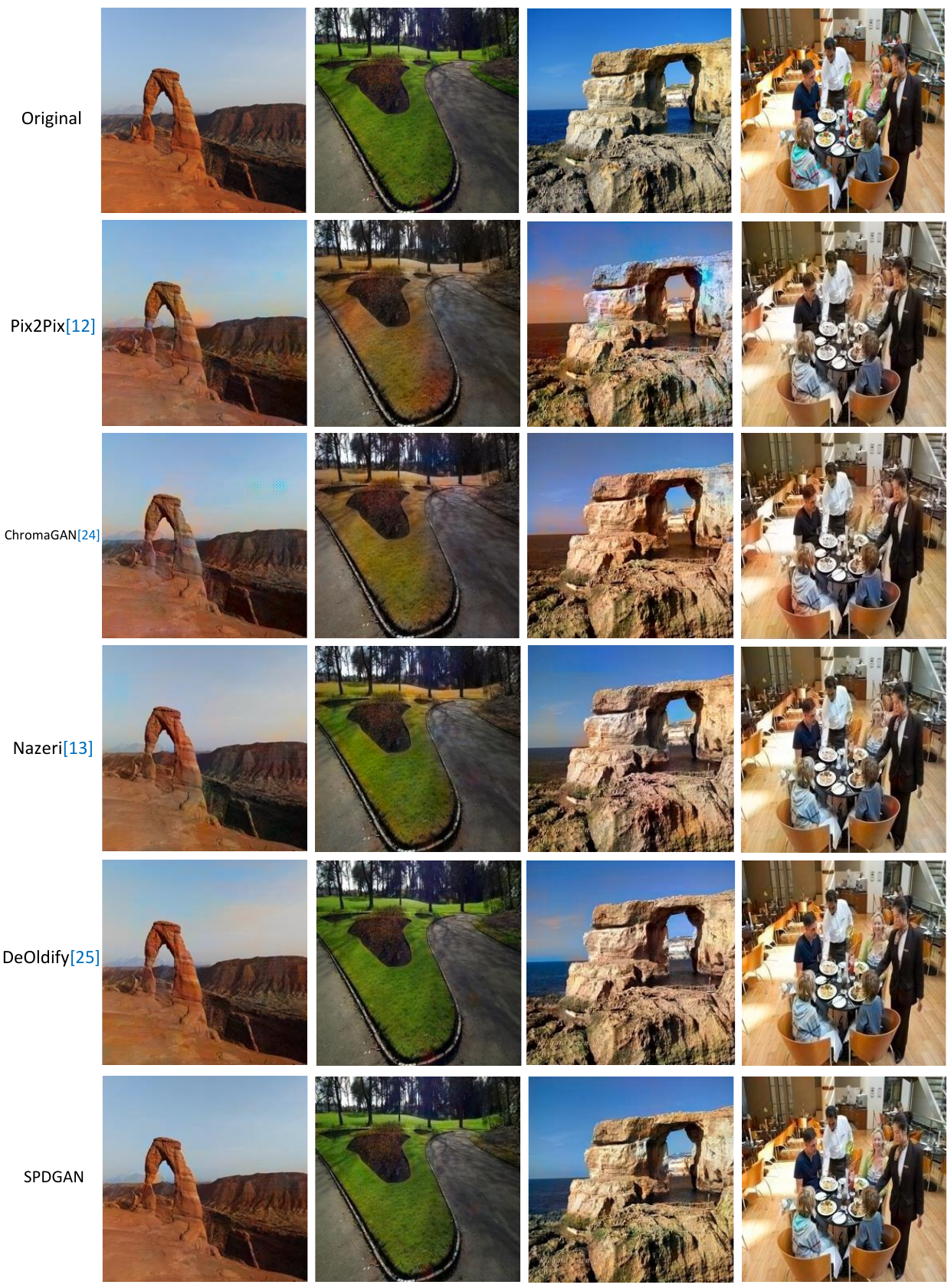}
\caption{Figure is better seen zoomed on the digital version of this document.  Qualitative comparisons of our proposed SPDGAN with state-of-the-art approaches on Places365 \cite{zhou2017places} dataset. Our SPDGAN is able to generate more realistic images with plausible color results. }
\label{fig:places365}
\end{figure*}

Figures \ref{fig:coco} and \ref{fig:places365} show the qualitative results of ou SPDGAN compared to comparison approaches. We tried to propose a set of colorful natural images, which covers a wide variety of content including animals, objects, and landscapes. We can observe that in most cases our SPDGAN achieves better results than the four state-of-the-art approaches in comparison. As we can see, due to texture similarity and scale variation, existing approaches produce colorized images with many mismatches for example in the first three columns, resulting in semantically wrong results (yellow grass, blue foot, uncaptured green color, etc.). With the help of the SPD discriminator and the color loss, the proposed model based on GAN produces plausible and semantically correct colorization results compared to state-of-the-art methods.

\subsubsection{Subjective user study}

It is known that metrics such as PSNR and SSIM can deviate substantially from human perceptual differences. Improved methods have been developed for image colorization assessment in general, such as FID and Colorfulness Score. Nevertheless, such metrics are not always appropriate for image colorization tasks, because one may obtain colorization results that still be perfectly plausible but are different from the ground truth. In order to make a fair comparison between our SPDGAN and state-of-the-art approaches, a user study is designed for evaluation. We invited 20 users in the range age 21-50 to participate in this user study including Ph.D. students and candidates without any image processing knowledge. We randomly select 70 images on a combined dataset including Places365 \cite{zhou2017places}, and Coco-Stuff \cite{caesar2018coco} from various categories of image contents. To avoid bias, colorization results are displayed in a random manner before showing it to the participants. The result of each approach is compared with the rest of the approaches. We save the total number of user preferences for all the images, and consider them as random variables. As observed in Table \ref{tabnaturalness}, 50.2\% of the images colorized using our SPDGAN obtained a significant preference by subjects, which compares favorably to 15.8\% obtained by Deoldify \cite{DeOldify}, 12.2\% by Nazeri \emph{et al.} \cite{nazeri2018image}, 11.6\% ChromaGAN \cite{vitoria2020chromagan}, 10.2\% Pix2Pix \cite{isola2017image}. Results demonstrate the distinct advantage on producing natural and plausible colorization by our SPDGAN.

\begin{table}[ht]
\begin{center}
\caption{Result of our user study. We evaluate the naturalness of colorization by the percentage of test images that are identified as real by users}
\begin{tabular}{llllllllllll}
    \hline\noalign{\smallskip}  
       Methods  & Naturalness\\ 
            \noalign{\smallskip}\hline\noalign{\smallskip}           
          DeOldify \cite{DeOldify}  & 15.8\%      \\
          Nazeri \emph{et al.} \cite{nazeri2018image} & 12.2\%    \\
          Pix2Pix \cite{isola2017image} & 10.2\%    \\

         ChromaGAN \cite{vitoria2020chromagan} & 11.6\%   \\
         SPDGAN (ours) & \textbf{50.2\%}     \\
        
    \hline      
 \end{tabular} 
 \label{tabnaturalness}
 \end{center}
 \end{table}

\begin{figure}[ht]
\centering
\includegraphics[width=13cm,height=6.5cm]{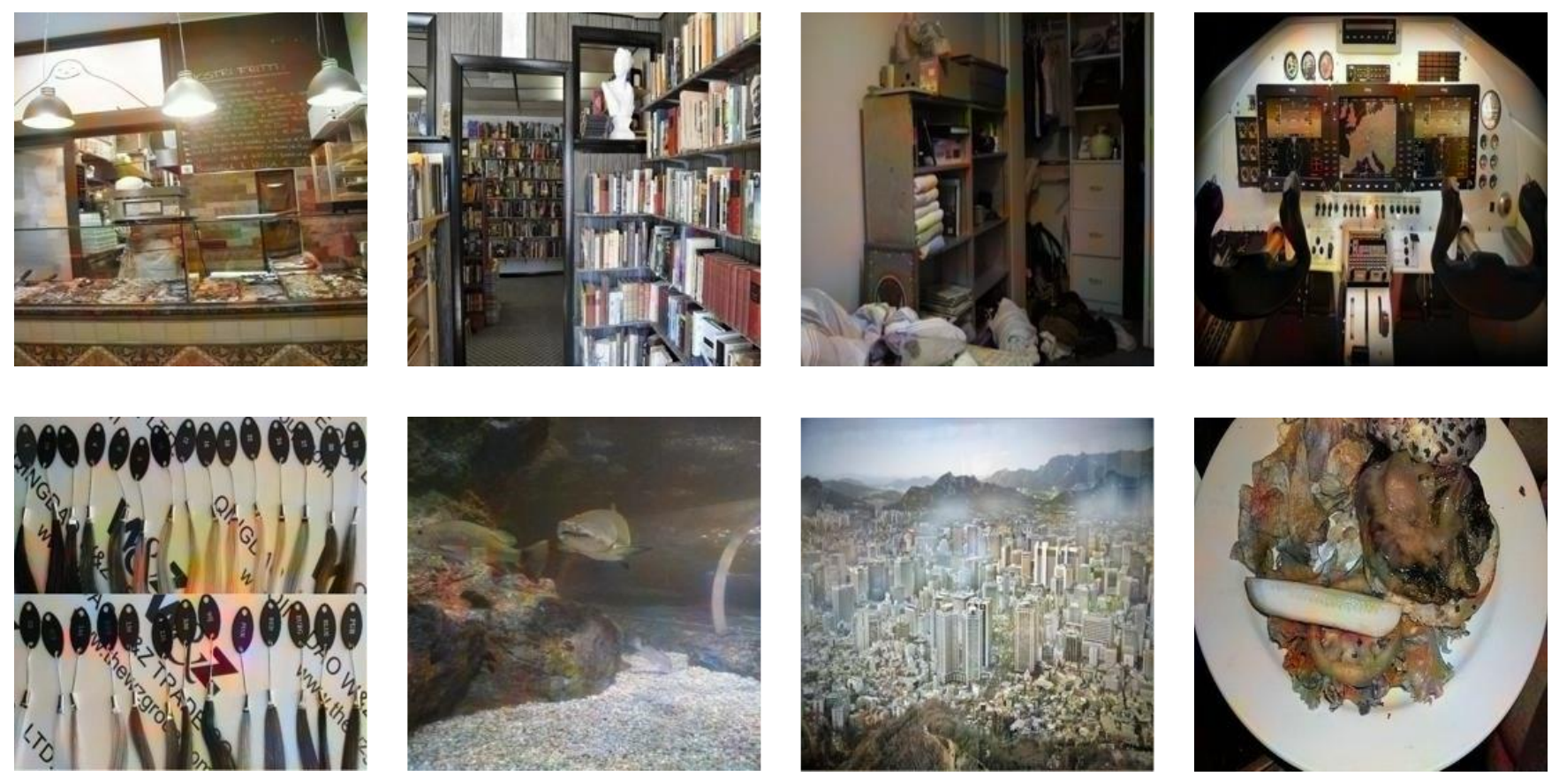}
\caption{Failure cases. Food, tiny objects, and underwater images are still very challenging.}
\label{fig:failures}
\end{figure}

\subsection{Failure Cases}

Our SPDGAN can produce realistic colorized images but
it is not perfect. There are still some common issues encountered by our model as well as other automatic colorization approaches. We present some of failure cases in Fig. \ref{fig:failures}. We believe that it is highly challenging to colorize different kinds of food and small objects. They are artificial and variable. In addition, our SPDGAN at this time cannot handle objects with unclear colorization and underwater. We believe that a finer semantic colorization with more image examples of these types of patterns (food, small objects, etc) will further enhance the colorization results.

\section{Conclusion}
\label{sec:conclu}

In this paper, we proposed a SPDGAN model that overcomes the limitation of automatic colorization. The proposed model takes advantage of GANs architecture and Riemann manifold learning. It employs a generator with two discriminators for the pixels and features domain. The generator is based on Resnet architecture that prevents losing image information across layers. The Image discriminator gets colored images from both generator and ground truth along with the grayscale image as the condition, and tries to distinguish which pair contains the true colored image. The SPD discriminator takes SPD matrices (usually used in the context of the
color transfer) constructed from VGG19 features and tries to decide which pair contains the true color distribution. In addition, to evaluate the difference in contrast, brightness, and major colors between images, we benefit from the advantages of non-adversarial color loss. Extensive experiments have been conducted in order to demonstrate the effect of different components of our SPDGAN. Through comparisons with state-of-the-art approaches on the prominent Places365 and COCO-Stuff datasets, it is shown that the proposed model generates realistic and plausible colorized images and overcomes effective literature approaches.

\section*{Declarations}

\subsection*{Funding}

\begin{itemize}
    \item The authors did not receive support from any organization for the submitted work.
    \item No funding was received to assist with the preparation of this manuscript.
    \item No funding was received for conducting this study.
    \item No funds, grants, or other support was received.
\end{itemize}

\subsection*{Competing interests}

\begin{itemize}
\item The authors have no relevant financial or non-financial interests to disclose.
\item The authors have no competing interests to declare that are relevant to the content of this article.
\item All authors certify that they have no affiliations with or involvement in any organization or entity with any financial interest or non-financial interest in the subject matter or materials discussed in this manuscript.
\item The authors have no financial or proprietary interests in any material discussed in this article.
\end{itemize}

\subsection*{Availability of data and materials}

\begin{itemize}
\item The datasets generated during and/or analyzed during the current study are available in [https://github.com/CSAILVision/places365] and [https://github.com/nightrome/cocostuff]
\end{itemize}

\bibliography{refs.bib}


\begin{thebibliography}{41}
\ifx \bisbn   \undefined \def \bisbn  #1{ISBN #1}\fi
\ifx \binits  \undefined \def \binits#1{#1}\fi
\ifx \bauthor  \undefined \def \bauthor#1{#1}\fi
\ifx \batitle  \undefined \def \batitle#1{#1}\fi
\ifx \bjtitle  \undefined \def \bjtitle#1{#1}\fi
\ifx \bvolume  \undefined \def \bvolume#1{\textbf{#1}}\fi
\ifx \byear  \undefined \def \byear#1{#1}\fi
\ifx \bissue  \undefined \def \bissue#1{#1}\fi
\ifx \bfpage  \undefined \def \bfpage#1{#1}\fi
\ifx \blpage  \undefined \def \blpage #1{#1}\fi
\ifx \burl  \undefined \def \burl#1{\textsf{#1}}\fi
\ifx \doiurl  \undefined \def \doiurl#1{\url{https://doi.org/#1}}\fi
\ifx \betal  \undefined \def \betal{\textit{et al.}}\fi
\ifx \binstitute  \undefined \def \binstitute#1{#1}\fi
\ifx \binstitutionaled  \undefined \def \binstitutionaled#1{#1}\fi
\ifx \bctitle  \undefined \def \bctitle#1{#1}\fi
\ifx \beditor  \undefined \def \beditor#1{#1}\fi
\ifx \bpublisher  \undefined \def \bpublisher#1{#1}\fi
\ifx \bbtitle  \undefined \def \bbtitle#1{#1}\fi
\ifx \bedition  \undefined \def \bedition#1{#1}\fi
\ifx \bseriesno  \undefined \def \bseriesno#1{#1}\fi
\ifx \blocation  \undefined \def \blocation#1{#1}\fi
\ifx \bsertitle  \undefined \def \bsertitle#1{#1}\fi
\ifx \bsnm \undefined \def \bsnm#1{#1}\fi
\ifx \bsuffix \undefined \def \bsuffix#1{#1}\fi
\ifx \bparticle \undefined \def \bparticle#1{#1}\fi
\ifx \barticle \undefined \def \barticle#1{#1}\fi
\bibcommenthead
\ifx \bconfdate \undefined \def \bconfdate #1{#1}\fi
\ifx \botherref \undefined \def \botherref #1{#1}\fi
\ifx \url \undefined \def \url#1{\textsf{#1}}\fi
\ifx \bchapter \undefined \def \bchapter#1{#1}\fi
\ifx \bbook \undefined \def \bbook#1{#1}\fi
\ifx \bcomment \undefined \def \bcomment#1{#1}\fi
\ifx \oauthor \undefined \def \oauthor#1{#1}\fi
\ifx \citeauthoryear \undefined \def \citeauthoryear#1{#1}\fi
\ifx \endbibitem  \undefined \def \endbibitem {}\fi
\ifx \bconflocation  \undefined \def \bconflocation#1{#1}\fi
\ifx \arxivurl  \undefined \def \arxivurl#1{\textsf{#1}}\fi
\csname PreBibitemsHook\endcsname

\bibitem{mourchid2016image}
\begin{barticle}
\bauthor{\bsnm{Mourchid}, \binits{Y.}},
\bauthor{\bsnm{El~Hassouni}, \binits{M.}},
\bauthor{\bsnm{Cherif}, \binits{H.}}:
\batitle{Image segmentation based on community detection approach}.
\bjtitle{International Journal of Computer Information Systems and Industrial Management Applications}
\bvolume{8},
\bfpage{10}--\blpage{10}
(\byear{2016})
\end{barticle}
\endbibitem

\bibitem{benallal2022new}
\begin{bchapter}
\bauthor{\bsnm{Benallal}, \binits{H.}},
\bauthor{\bsnm{Mourchid}, \binits{Y.}},
\bauthor{\bsnm{Abouelaziz}, \binits{I.}},
\bauthor{\bsnm{Alfalou}, \binits{A.}},
\bauthor{\bsnm{Tairi}, \binits{H.}},
\bauthor{\bsnm{Riffi}, \binits{J.}},
\bauthor{\bsnm{El~Hassouni}, \binits{M.}}:
\bctitle{A new approach for removing point cloud outliers using box plot}.
In: \bbtitle{Pattern Recognition and Tracking XXXIII},
vol. \bseriesno{12101},
pp. \bfpage{63}--\blpage{69}
(\byear{2022}).
\bcomment{SPIE}
\end{bchapter}
\endbibitem

\bibitem{mourchid2021automatic}
\begin{bchapter}
\bauthor{\bsnm{Mourchid}, \binits{Y.}},
\bauthor{\bsnm{Donias}, \binits{M.}},
\bauthor{\bsnm{Berthoumieu}, \binits{Y.}}:
\bctitle{Automatic image colorization based on multi-discriminators generative adversarial networks}.
In: \bbtitle{2020 28th European Signal Processing Conference (EUSIPCO)},
pp. \bfpage{1532}--\blpage{1536}
(\byear{2021}).
\bcomment{IEEE}
\end{bchapter}
\endbibitem

\bibitem{mourchid2023mr}
\begin{bchapter}
\bauthor{\bsnm{Mourchid}, \binits{Y.}},
\bauthor{\bsnm{Slama}, \binits{R.}}:
\bctitle{Mr-stgn: Multi-residual spatio temporal graph network using attention fusion for patient action assessment}.
In: \bbtitle{2023 IEEE 25th International Workshop on Multimedia Signal Processing (MMSP)},
pp. \bfpage{1}--\blpage{6}
(\byear{2023}).
\bcomment{IEEE}
\end{bchapter}
\endbibitem

\bibitem{levin2004colorization}
\begin{bchapter}
\bauthor{\bsnm{Levin}, \binits{A.}},
\bauthor{\bsnm{Lischinski}, \binits{D.}},
\bauthor{\bsnm{Weiss}, \binits{Y.}}:
\bctitle{Colorization using optimization}.
In: \bbtitle{ACM Transactions on Graphics (tog)},
vol. \bseriesno{23},
pp. \bfpage{689}--\blpage{694}
(\byear{2004}).
\bcomment{ACM}
\end{bchapter}
\endbibitem

\bibitem{huang2005adaptive}
\begin{bchapter}
\bauthor{\bsnm{Huang}, \binits{Y.-C.}},
\bauthor{\bsnm{Tung}, \binits{Y.-S.}},
\bauthor{\bsnm{Chen}, \binits{J.-C.}},
\bauthor{\bsnm{Wang}, \binits{S.-W.}},
\bauthor{\bsnm{Wu}, \binits{J.-L.}}:
\bctitle{An adaptive edge detection based colorization algorithm and its applications}.
In: \bbtitle{Proceedings of the 13th Annual ACM International Conference on Multimedia},
pp. \bfpage{351}--\blpage{354}
(\byear{2005})
\end{bchapter}
\endbibitem

\bibitem{luan2007natural}
\begin{bchapter}
\bauthor{\bsnm{Luan}, \binits{Q.}},
\bauthor{\bsnm{Wen}, \binits{F.}},
\bauthor{\bsnm{Cohen-Or}, \binits{D.}},
\bauthor{\bsnm{Liang}, \binits{L.}},
\bauthor{\bsnm{Xu}, \binits{Y.-Q.}},
\bauthor{\bsnm{Shum}, \binits{H.-Y.}}:
\bctitle{Natural image colorization}.
In: \bbtitle{Proceedings of the 18th Eurographics Conference on Rendering Techniques},
pp. \bfpage{309}--\blpage{320}
(\byear{2007})
\end{bchapter}
\endbibitem

\bibitem{welsh2002transferring}
\begin{bchapter}
\bauthor{\bsnm{Welsh}, \binits{T.}},
\bauthor{\bsnm{Ashikhmin}, \binits{M.}},
\bauthor{\bsnm{Mueller}, \binits{K.}}:
\bctitle{Transferring color to greyscale images}.
In: \bbtitle{Proceedings of the 29th Annual Conference on Computer Graphics and Interactive Techniques},
pp. \bfpage{277}--\blpage{280}
(\byear{2002})
\end{bchapter}
\endbibitem

\bibitem{chia2011semantic}
\begin{barticle}
\bauthor{\bsnm{Chia}, \binits{A.Y.-S.}},
\bauthor{\bsnm{Zhuo}, \binits{S.}},
\bauthor{\bsnm{Gupta}, \binits{R.K.}},
\bauthor{\bsnm{Tai}, \binits{Y.-W.}},
\bauthor{\bsnm{Cho}, \binits{S.-Y.}},
\bauthor{\bsnm{Tan}, \binits{P.}},
\bauthor{\bsnm{Lin}, \binits{S.}}:
\batitle{Semantic colorization with internet images}.
\bjtitle{ACM Transactions on Graphics (TOG)}
\bvolume{30}(\bissue{6}),
\bfpage{1}--\blpage{8}
(\byear{2011})
\end{barticle}
\endbibitem

\bibitem{liu2008intrinsic}
\begin{bchapter}
\bauthor{\bsnm{Liu}, \binits{X.}},
\bauthor{\bsnm{Wan}, \binits{L.}},
\bauthor{\bsnm{Qu}, \binits{Y.}},
\bauthor{\bsnm{Wong}, \binits{T.-T.}},
\bauthor{\bsnm{Lin}, \binits{S.}},
\bauthor{\bsnm{Leung}, \binits{C.-S.}},
\bauthor{\bsnm{Heng}, \binits{P.-A.}}:
\bctitle{Intrinsic colorization}.
In: \bbtitle{ACM SIGGRAPH Asia 2008 Papers},
pp. \bfpage{1}--\blpage{9}
(\byear{2008})
\end{bchapter}
\endbibitem

\bibitem{deshpande2015learning}
\begin{bchapter}
\bauthor{\bsnm{Deshpande}, \binits{A.}},
\bauthor{\bsnm{Rock}, \binits{J.}},
\bauthor{\bsnm{Forsyth}, \binits{D.}}:
\bctitle{Learning large-scale automatic image colorization}.
In: \bbtitle{Proceedings of the IEEE International Conference on Computer Vision},
pp. \bfpage{567}--\blpage{575}
(\byear{2015})
\end{bchapter}
\endbibitem

\bibitem{cheng2015deep}
\begin{bchapter}
\bauthor{\bsnm{Cheng}, \binits{Z.}},
\bauthor{\bsnm{Yang}, \binits{Q.}},
\bauthor{\bsnm{Sheng}, \binits{B.}}:
\bctitle{Deep colorization}.
In: \bbtitle{Proceedings of the IEEE International Conference on Computer Vision},
pp. \bfpage{415}--\blpage{423}
(\byear{2015})
\end{bchapter}
\endbibitem

\bibitem{zhang2016colorful}
\begin{bchapter}
\bauthor{\bsnm{Zhang}, \binits{R.}},
\bauthor{\bsnm{Isola}, \binits{P.}},
\bauthor{\bsnm{Efros}, \binits{A.A.}}:
\bctitle{Colorful image colorization}.
In: \bbtitle{European Conference on Computer Vision},
pp. \bfpage{649}--\blpage{666}
(\byear{2016}).
\bcomment{Springer}
\end{bchapter}
\endbibitem

\bibitem{larsson2016learning}
\begin{bchapter}
\bauthor{\bsnm{Larsson}, \binits{G.}},
\bauthor{\bsnm{Maire}, \binits{M.}},
\bauthor{\bsnm{Shakhnarovich}, \binits{G.}}:
\bctitle{Learning representations for automatic colorization}.
In: \bbtitle{European Conference on Computer Vision},
pp. \bfpage{577}--\blpage{593}
(\byear{2016}).
\bcomment{Springer}
\end{bchapter}
\endbibitem

\bibitem{zhang2017real}
\begin{barticle}
\bauthor{\bsnm{Zhang}, \binits{R.Y.}},
\bauthor{\bsnm{Zhu}, \binits{J.Y.}},
\bauthor{\bsnm{Isola}, \binits{P.}},
\bauthor{\bsnm{Geng}, \binits{X.}},
\bauthor{\bsnm{Lin}, \binits{A.S.}},
\bauthor{\bsnm{Yu}, \binits{T.}},
\bauthor{\bsnm{Efros}, \binits{A.A.}}:
\batitle{Real-time user-guided image colorization with learned deep priors}.
\bjtitle{ACM Transactions on Graphics}
\bvolume{36}(\bissue{4}),
\bfpage{119}
(\byear{2017})
\end{barticle}
\endbibitem

\bibitem{xia2018scene}
\begin{botherref}
\oauthor{\bsnm{Xia}, \binits{Y.}},
\oauthor{\bsnm{Qu}, \binits{S.}},
\oauthor{\bsnm{Wan}, \binits{S.}}:
Scene guided colorization using neural networks.
Neural Computing and Applications,
1--14
(2018)
\end{botherref}
\endbibitem

\bibitem{isola2017image}
\begin{bchapter}
\bauthor{\bsnm{Isola}, \binits{P.}},
\bauthor{\bsnm{Zhu}, \binits{J.-Y.}},
\bauthor{\bsnm{Zhou}, \binits{T.}},
\bauthor{\bsnm{Efros}, \binits{A.A.}}:
\bctitle{Image-to-image translation with conditional adversarial networks}.
In: \bbtitle{Proceedings of the IEEE Conference on Computer Vision and Pattern Recognition},
pp. \bfpage{1125}--\blpage{1134}
(\byear{2017})
\end{bchapter}
\endbibitem

\bibitem{nazeri2018image}
\begin{bchapter}
\bauthor{\bsnm{Nazeri}, \binits{K.}},
\bauthor{\bsnm{Ng}, \binits{E.}},
\bauthor{\bsnm{Ebrahimi}, \binits{M.}}:
\bctitle{Image colorization using generative adversarial networks}.
In: \bbtitle{International Conference on Articulated Motion and Deformable Objects},
pp. \bfpage{85}--\blpage{94}
(\byear{2018}).
\bcomment{Springer}
\end{bchapter}
\endbibitem

\bibitem{vitoria2020chromagan}
\begin{bchapter}
\bauthor{\bsnm{Vitoria}, \binits{P.}},
\bauthor{\bsnm{Raad}, \binits{L.}},
\bauthor{\bsnm{Ballester}, \binits{C.}}:
\bctitle{Chromagan: Adversarial picture colorization with semantic class distribution}.
In: \bbtitle{Proceedings of the IEEE/CVF Winter Conference on Applications of Computer Vision},
pp. \bfpage{2445}--\blpage{2454}
(\byear{2020})
\end{bchapter}
\endbibitem

\bibitem{mourchid2020dual}
\begin{bchapter}
\bauthor{\bsnm{Mourchid}, \binits{Y.}},
\bauthor{\bsnm{Donias}, \binits{M.}},
\bauthor{\bsnm{Berthoumieu}, \binits{Y.}}:
\bctitle{Dual color-image discriminators adversarial networks for generating artificial-sar colorized images from sentinel-1 images}.
In: \bbtitle{Machine Learning for Earth Observation Workshop (ECML/PKDD 2020)}
(\byear{2020})
\end{bchapter}
\endbibitem

\bibitem{charpiat2008automatic}
\begin{bchapter}
\bauthor{\bsnm{Charpiat}, \binits{G.}},
\bauthor{\bsnm{Hofmann}, \binits{M.}},
\bauthor{\bsnm{Sch{\"o}lkopf}, \binits{B.}}:
\bctitle{Automatic image colorization via multimodal predictions}.
In: \bbtitle{European Conference on Computer Vision},
pp. \bfpage{126}--\blpage{139}
(\byear{2008}).
\bcomment{Springer}
\end{bchapter}
\endbibitem

\bibitem{he2015delving}
\begin{bchapter}
\bauthor{\bsnm{He}, \binits{K.}},
\bauthor{\bsnm{Zhang}, \binits{X.}},
\bauthor{\bsnm{Ren}, \binits{S.}},
\bauthor{\bsnm{Sun}, \binits{J.}}:
\bctitle{Delving deep into rectifiers: Surpassing human-level performance on imagenet classification}.
In: \bbtitle{Proceedings of the IEEE International Conference on Computer Vision},
pp. \bfpage{1026}--\blpage{1034}
(\byear{2015})
\end{bchapter}
\endbibitem

\bibitem{goodfellow2014generative}
\begin{bchapter}
\bauthor{\bsnm{Goodfellow}, \binits{I.}},
\bauthor{\bsnm{Pouget-Abadie}, \binits{J.}},
\bauthor{\bsnm{Mirza}, \binits{M.}},
\bauthor{\bsnm{Xu}, \binits{B.}},
\bauthor{\bsnm{Warde-Farley}, \binits{D.}},
\bauthor{\bsnm{Ozair}, \binits{S.}},
\bauthor{\bsnm{Courville}, \binits{A.}},
\bauthor{\bsnm{Bengio}, \binits{Y.}}:
\bctitle{Generative adversarial nets}.
In: \bbtitle{Advances in Neural Information Processing Systems},
pp. \bfpage{2672}--\blpage{2680}
(\byear{2014})
\end{bchapter}
\endbibitem

\bibitem{gatys2016image}
\begin{bchapter}
\bauthor{\bsnm{Gatys}, \binits{L.A.}},
\bauthor{\bsnm{Ecker}, \binits{A.S.}},
\bauthor{\bsnm{Bethge}, \binits{M.}}:
\bctitle{Image style transfer using convolutional neural networks}.
In: \bbtitle{Proceedings of the IEEE Conference on Computer Vision and Pattern Recognition},
pp. \bfpage{2414}--\blpage{2423}
(\byear{2016})
\end{bchapter}
\endbibitem

\bibitem{johnson2016perceptual}
\begin{bchapter}
\bauthor{\bsnm{Johnson}, \binits{J.}},
\bauthor{\bsnm{Alahi}, \binits{A.}},
\bauthor{\bsnm{Fei-Fei}, \binits{L.}}:
\bctitle{Perceptual losses for real-time style transfer and super-resolution}.
In: \bbtitle{European Conference on Computer Vision},
pp. \bfpage{694}--\blpage{711}
(\byear{2016}).
\bcomment{Springer}
\end{bchapter}
\endbibitem

\bibitem{balas2006texture}
\begin{barticle}
\bauthor{\bsnm{Balas}, \binits{B.J.}}:
\batitle{Texture synthesis and perception: Using computational models to study texture representations in the human visual system}.
\bjtitle{Vision Research}
\bvolume{46}(\bissue{3}),
\bfpage{299}--\blpage{309}
(\byear{2006})
\end{barticle}
\endbibitem

\bibitem{zhen2021flow}
\begin{bchapter}
\bauthor{\bsnm{Zhen}, \binits{X.}},
\bauthor{\bsnm{Chakraborty}, \binits{R.}},
\bauthor{\bsnm{Yang}, \binits{L.}},
\bauthor{\bsnm{Singh}, \binits{V.}}:
\bctitle{Flow-based generative models for learning manifold to manifold mappings}.
In: \bbtitle{Proceedings of the AAAI Conference on Artificial Intelligence},
vol. \bseriesno{35},
pp. \bfpage{11042}--\blpage{11052}
(\byear{2021})
\end{bchapter}
\endbibitem

\bibitem{huang2017riemannian}
\begin{bchapter}
\bauthor{\bsnm{Huang}, \binits{Z.}},
\bauthor{\bsnm{Van~Gool}, \binits{L.}}:
\bctitle{A riemannian network for spd matrix learning}.
In: \bbtitle{Thirty-First AAAI Conference on Artificial Intelligence}
(\byear{2017})
\end{bchapter}
\endbibitem

\bibitem{wang2022image}
\begin{barticle}
\bauthor{\bsnm{Wang}, \binits{N.}},
\bauthor{\bsnm{Chen}, \binits{G.-D.}},
\bauthor{\bsnm{Tian}, \binits{Y.}}:
\batitle{Image colorization algorithm based on deep learning}.
\bjtitle{Symmetry}
\bvolume{14}(\bissue{11}),
\bfpage{2295}
(\byear{2022})
\end{barticle}
\endbibitem

\bibitem{radford2015unsupervised}
\begin{botherref}
\oauthor{\bsnm{Radford}, \binits{A.}},
\oauthor{\bsnm{Metz}, \binits{L.}},
\oauthor{\bsnm{Chintala}, \binits{S.}}:
Unsupervised representation learning with deep convolutional generative adversarial networks.
arXiv preprint arXiv:1511.06434
(2015)
\end{botherref}
\endbibitem

\bibitem{xiao2019single}
\begin{bchapter}
\bauthor{\bsnm{Xiao}, \binits{Y.}},
\bauthor{\bsnm{Jiang}, \binits{A.}},
\bauthor{\bsnm{Liu}, \binits{C.}},
\bauthor{\bsnm{Wang}, \binits{M.}}:
\bctitle{Single image colorization via modified cyclegan}.
In: \bbtitle{IEEE International Conference on Image Processing (ICIP)},
pp. \bfpage{3247}--\blpage{3251}
(\byear{2019})
\end{bchapter}
\endbibitem

\bibitem{DeOldify}
\begin{botherref}
\oauthor{\bsnm{Antic}, \binits{J.}}:
Deoldify.
[Online]. Available: https://github.com/jantic/
(2019)
\end{botherref}
\endbibitem

\bibitem{li2017demystifying}
\begin{botherref}
\oauthor{\bsnm{Li}, \binits{Y.}},
\oauthor{\bsnm{Wang}, \binits{N.}},
\oauthor{\bsnm{Liu}, \binits{J.}},
\oauthor{\bsnm{Hou}, \binits{X.}}:
Demystifying neural style transfer.
arXiv preprint arXiv:1701.01036
(2017)
\end{botherref}
\endbibitem

\bibitem{arsigny2007geometric}
\begin{barticle}
\bauthor{\bsnm{Arsigny}, \binits{V.}},
\bauthor{\bsnm{Fillard}, \binits{P.}},
\bauthor{\bsnm{Pennec}, \binits{X.}},
\bauthor{\bsnm{Ayache}, \binits{N.}}:
\batitle{Geometric means in a novel vector space structure on symmetric positive-definite matrices}.
\bjtitle{SIAM Journal on Matrix Analysis and Applications}
\bvolume{29}(\bissue{1}),
\bfpage{328}--\blpage{347}
(\byear{2007})
\end{barticle}
\endbibitem

\bibitem{ioffe2015batch}
\begin{bchapter}
\bauthor{\bsnm{Ioffe}, \binits{S.}},
\bauthor{\bsnm{Szegedy}, \binits{C.}}:
\bctitle{Batch normalization: Accelerating deep network training by reducing internal covariate shift}.
In: \bbtitle{International Conference on Machine Learning},
pp. \bfpage{448}--\blpage{456}
(\byear{2015}).
\bcomment{PMLR}
\end{bchapter}
\endbibitem

\bibitem{miyato2018spectral}
\begin{botherref}
\oauthor{\bsnm{Miyato}, \binits{T.}},
\oauthor{\bsnm{Kataoka}, \binits{T.}},
\oauthor{\bsnm{Koyama}, \binits{M.}},
\oauthor{\bsnm{Yoshida}, \binits{Y.}}:
Spectral normalization for generative adversarial networks.
arXiv preprint arXiv:1802.05957
(2018)
\end{botherref}
\endbibitem

\bibitem{ulyanov2016instance}
\begin{botherref}
\oauthor{\bsnm{Ulyanov}, \binits{D.}},
\oauthor{\bsnm{Vedaldi}, \binits{A.}},
\oauthor{\bsnm{Lempitsky}, \binits{V.}}:
Instance normalization: The missing ingredient for fast stylization.
arXiv preprint arXiv:1607.08022
(2016)
\end{botherref}
\endbibitem

\bibitem{zhou2017places}
\begin{barticle}
\bauthor{\bsnm{Zhou}, \binits{B.}},
\bauthor{\bsnm{Lapedriza}, \binits{A.}},
\bauthor{\bsnm{Khosla}, \binits{A.}},
\bauthor{\bsnm{Oliva}, \binits{A.}},
\bauthor{\bsnm{Torralba}, \binits{A.}}:
\batitle{Places: A 10 million image database for scene recognition}.
\bjtitle{IEEE transactions on pattern analysis and machine intelligence}
\bvolume{40}(\bissue{6}),
\bfpage{1452}--\blpage{1464}
(\byear{2017})
\end{barticle}
\endbibitem

\bibitem{caesar2018coco}
\begin{bchapter}
\bauthor{\bsnm{Caesar}, \binits{H.}},
\bauthor{\bsnm{Uijlings}, \binits{J.}},
\bauthor{\bsnm{Ferrari}, \binits{V.}}:
\bctitle{Coco-stuff: Thing and stuff classes in context}.
In: \bbtitle{Proceedings of the IEEE Conference on Computer Vision and Pattern Recognition},
pp. \bfpage{1209}--\blpage{1218}
(\byear{2018})
\end{bchapter}
\endbibitem

\bibitem{heusel2017gans}
\begin{botherref}
\oauthor{\bsnm{Heusel}, \binits{M.}},
\oauthor{\bsnm{Ramsauer}, \binits{H.}},
\oauthor{\bsnm{Unterthiner}, \binits{T.}},
\oauthor{\bsnm{Nessler}, \binits{B.}},
\oauthor{\bsnm{Hochreiter}, \binits{S.}}:
Gans trained by a two time-scale update rule converge to a local nash equilibrium.
Advances in neural information processing systems
\textbf{30}
(2017)
\end{botherref}
\endbibitem

\bibitem{hasler2003measuring}
\begin{bchapter}
\bauthor{\bsnm{Hasler}, \binits{D.}},
\bauthor{\bsnm{Suesstrunk}, \binits{S.E.}}:
\bctitle{Measuring colorfulness in natural images}.
In: \bbtitle{Human Vision and Electronic Imaging VIII},
vol. \bseriesno{5007},
pp. \bfpage{87}--\blpage{95}
(\byear{2003}).
\bcomment{SPIE}
\end{bchapter}
\endbibitem

\end{thebibliography}

\end{document}